\documentclass[Afour,sageh,times]{sagej}

\usepackage{moreverb,url}

\usepackage[colorlinks,bookmarksopen,bookmarksnumbered,citecolor=red,urlcolor=red]{hyperref}
\usepackage{multirow}
\usepackage{graphicx}
\usepackage{makecell}
\usepackage{rotating}

\usepackage{tabularx}
\usepackage{booktabs} 

\usepackage{siunitx}
\usepackage{subcaption}
\usepackage[markup=underlined, commandnameprefix=ifneeded]{changes}

\newcommand\BibTeX{{\rmfamily B\kern-.05em \textsc{i\kern-.025em b}\kern-.08em
T\kern-.1667em\lower.7ex\hbox{E}\kern-.125emX}}

\def\volumeyear{2016}

\begin{document}

\runninghead{Zhan et al.}

\title{AgriLiRa4D: A Multi-Sensor UAV Dataset for Robust SLAM in Challenging Agricultural Fields}

\author{Zhihao Zhan\affilnum{1,2}, Yuhang Ming\affilnum{3}, Shaobin Li\affilnum{2}, and Jie Yuan\affilnum{1}}

\affiliation{\affilnum{1}School of Electronic Science and Engineering, Nanjing University, Nanjing, 210023, China \\
\affilnum{2}TopXGun Robotics, Nanjing, 211100, China \\
\affilnum{3}School of Computer Science, Hangzhou Dianzi University, Hangzhou, 310018, China}

\corrauth{
Yuhang Ming, 
School of Computer Science,
Hangzhou Dianzi University,
Hangzhou, 310018, China. \\
Jie Yuan, 
School of Electronic Science and Engineering, 
Nanjing University, 
Nanjing, 210023, China.}

\email{yuhang.ming@hdu.edu.cn, yuanjie@nju.edu.cn}

\begin{abstract}
Multi-sensor simultaneous localization and mapping (SLAM) is essential for unmanned aerial vehicles (UAVs) performing agricultural tasks such as spraying, surveying, and inspection. 
However, real-world, multi-modal agricultural UAV datasets that enable research on robust operation remain scarce. 
To address this gap, we present AgriLiRa4D, a multi-modal UAV dataset designed for challenging outdoor agricultural environments. 
AgriLiRa4D spans three representative farmland types—flat, hilly, and terraced—and includes both boundary and coverage operation modes, resulting in six flight sequence groups.
The dataset provides high-accuracy ground-truth trajectories from a fiber-optic inertial navigation system with real-time kinematic capability (FINS-RTK), along with synchronized measurements from a 3D lidar, a 4D radar, and an inertial measurement unit (IMU), accompanied by complete intrinsic and extrinsic calibrations.
Leveraging its comprehensive sensor suite and diverse real-world scenarios, AgriLiRa4D supports diverse SLAM and localization studies and enables rigorous robustness evaluation against low-texture crops, repetitive patterns, dynamic vegetation, and other challenges of real agricultural environments. 
To further demonstrate its utility, we benchmark four state-of-the-art multi-sensor SLAM algorithms across different sensor combinations, highlighting the difficulty of the proposed sequences and the challenges of effectively exploiting complementary sensing modalities for reliable UAV localization.
By filling a critical gap in agricultural SLAM datasets, AgriLiRa4D provides a valuable benchmark for the research community and contributes to advancing autonomous navigation technologies for agricultural UAVs.
The dataset can be downloaded from: \textcolor{red}{https://zhan994.github.io/AgriLiRa4D}.
\end{abstract}


\maketitle

\section{Introduction} \label{sec:intro} 
The rapid advancement of agricultural technology has positioned unmanned aerial vehicles (UAVs) as essential platforms for precision farming tasks such as crop monitoring, pesticide spraying, and large-scale farmland surveying~\citep{radoglou2020compilation, cheng2023precision, he2025multisensorfusionapproachrapid, wang2025high}.
To operate safely and efficiently in these complex outdoor environments, agricultural UAVs require robust and accurate localization as a fundamental capability for autonomous navigation and task execution.

Although global navigation satellite system (GNSS) positioning is widely used in agricultural applications, its performance often degrades severely due to vegetation occlusion, atmospheric disturbances, and multipath effects in structured farmlands~\citep{pini2020experimental}. 
As a result, GNSS alone is insufficient for reliable localization, particularly when UAVs operate near crop canopies or close to the terrain. 
These limitations motivate the adoption of simultaneous localization and mapping (SLAM)-based approaches, which can provide continuous pose estimation when GNSS becomes unreliable.

However, conventional visual SLAM systems face their own challenges in agricultural environments. 
Dynamic illumination, repetitive and texture-sparse crop patterns, and low-light or low-temperature operating conditions—commonly encountered during pesticide spraying—reduce the availability of stable visual features. 
In addition, high-speed UAV flight induces motion blur, while strong downwash airflow from heavy payloads causes vegetation motion, further destabilizing feature tracking and leading to frequent failures~\citep{cadena2017past, mur2017orb}.
These factors collectively limit the robustness of pure visual SLAM, underscoring the need for multi-sensor SLAM systems that integrate lidar, radar, inertial sensing, and other modalities to achieve reliable localization in challenging agricultural settings.

Recent advances in multi-modal sensor fusion have demonstrated promising potential for enhancing SLAM robustness in challenging environments~\citep{zuo2019lic,  shan2020lio, 10757429}. 
Lidar sensors provide accurate and illumination-invariant geometric measurements, while emerging 4D radar technology offers complementary advantages including weather robustness and direct velocity estimation~\citep{zhang20234dradarslam, zhuang20234d}. 
Fusing these ranging modalities with inertial measurement units (IMUs) therefore offers a promising pathway to overcoming the inherent limitations of GNSS- and vision-based systems in agricultural UAV applications.

Despite the theoretical advantages of multi-modal sensing, the development and evaluation of robust agricultural UAV SLAM systems remain constrained by the lack of specialized datasets that comprehensively capture the operational diversity of agricultural missions. 
Existing datasets predominantly focus on urban or indoor scenarios~\citep{burri2016ijrr-euroc, majdik2017ijrr-zurichmav, nguyen2022ijrr-ntuviral}, and only a few extending to semi-natural settings such as islands or rural towns~\citep{li2024ijrr-marslvig}. 
However, these semi-structured environments still differ substantially from real farmland, where unstructured terrain, heterogeneous vegetation, and large-scale outdoor operations introduce fundamentally different sensing and localization challenges. 
Agricultural UAVs must operate across a wide range of tasks, from low-altitude crop inspection to high-speed field mapping, each characterized by distinct motion dynamics and complex environmental interactions that critically affect SLAM robustness. 
These limitations highlight the need for a dedicated agricultural UAV dataset that systematically represents the sensing, motion, and environmental diversity inherent to real-world farming operations.

To address this critical gap, this paper introduces a novel multi-modal dataset specifically developed for agricultural UAV SLAM research, incorporating lidar, 4D radar, and IMU measurements collected using an agricultural UAV platform. 
Our dataset features high-precision ground truth trajectories obtained from a fiber-optic inertial navigation system (FINS) module with a built-in real-time kinematic (RTK) receiver (hereafter denoted as FINS-RTK), ensuring centimeter-level position accuracy and high-fidelity orientation references. 
This work aims to advance the development of robust SLAM systems tailored for agricultural autonomous operations. 
Our contributions are summarized as follows: 

\begin{enumerate}
    \item \textbf{AgriLiRa4D Dataset:} 
    We introduce a large-scale agricultural UAV dataset that fills the critical gap between the growing need for autonomous UAV operation in agriculture and the scarcity of real-world benchmarks. 
    It covers diverse terrain types, motion dynamics, and flight altitudes, and provides synchronized lidar, 4D radar, and IMU measurements with centimeter-level position and high-precision orientation ground truth from a FINS-RTK system.
    \item \textbf{SLAM Benchmark:} 
    We perform comprehensive evaluations of representative multi-sensor fusion SLAM algorithms (lidar–inertial, radar–inertial, and radar–lidar–inertial) under varied agricultural conditions, providing quantitative insights into their localization accuracy, robustness, and adaptability to complex outdoor environments.
    \item \textbf{Multi-Modal Fusion Analysis:} 
    We analyze the performance of lidar–inertial, radar–inertial, and radar–lidar–inertial methods across diverse agricultural scenarios. 
    The results reveal modality-dependent strengths and limitations and identify environmental conditions under which radar measurements may provide useful complementary information or remain difficult to exploit effectively.
\end{enumerate}

The remainder of this paper is organized as follows. 
Section~\ref{sec:related} provides a detailed comparison between our dataset and existing benchmarks. 
Section~\ref{sec:system} describes the UAV platform and sensing configuration used for data collection. 
Section~\ref{sec:dataset} presents the characteristics of the dataset, highlighting the diverse agricultural scenarios and motion patterns. 
Section~\ref{sec:exp} reports experimental evaluations and analysis of three categories of state-of-the-art SLAM methods. 
Finally, Section~\ref{sec:conclusion} concludes the paper and outlines potential directions for future work.


\section{Related Work} \label{sec:related}

\begin{sidewaystable*}[!p]
\centering
\caption{\textbf{Comparison of representative datasets.}
AgriLiRa4D is distinguished by its combination of lidar-4D radar-IMU sensing, FINS-RTK trajectory ground truth, and agricultural UAV flights across flat, hilly, and terraced farmland, enabling the evaluation of different sensing configurations under diverse agricultural terrain conditions.
}
\label{tab:datasets}

\small
\setlength{\tabcolsep}{8pt}
\renewcommand{\arraystretch}{1.1}

\scalebox{0.9}{
\begin{tabularx}{\textwidth}{
    p{0.5cm} 
    p{2cm}   
    p{4.2cm} 
    cccc 
    p{3cm}   
    p{3cm} 
    X 
}
\toprule
& \textbf{Platform} & \textbf{Dataset (Venue)} & \textbf{Lidar} & \textbf{4D Radar} & \textbf{IMU} & \textbf{Camera} &  \textbf{Ground Truth} & \textbf{Environment} & \textbf{Primary Task} \\
\midrule

\multirow{9}{*}{\rotatebox{90}{\parbox{8cm}{\centering\textbf{Multi-Sensor UAV-included Datasets}}}}
& Aerial &
\begin{tabular}[t]{@{}l@{}}
EuRoC (IJRR)\\
\cite{burri2016ijrr-euroc}
\end{tabular} & 
-- & -- & $\checkmark$ & $\checkmark$ &
Laser; MoCap & 
Indoor lab 
& Visual-Inertial SLAM \\

& Aerial &
\begin{tabular}[t]{@{}l@{}}
Zurich Urban MAV (IJRR)\\
\cite{majdik2017ijrr-zurichmav}
\end{tabular} & 
-- & -- & $\checkmark$ & $\checkmark$ &
Photogrammetric reconstruction &
Urban 
& SLAM; Reconstruction \\

& Aerial &
\begin{tabular}[t]{@{}l@{}}
UPenn Fast Flight (RAL)\\
\cite{sun2018ral-upennfastflight}
\end{tabular} &
-- & -- & $\checkmark$ & $\checkmark$ &
GPS &
Warehouse; Woodland 
& Visual-Inertial SLAM \\

& Aerial &
\begin{tabular}[t]{@{}l@{}}
UZH-FPV (ICRA)\\
\cite{delmerio2019icra-uzhfpv}
\end{tabular} &
-- & -- & $\checkmark$ & $\checkmark$ &
Laser &
Hanger; Grassland 
& Drone racing \\

& Aerial &
\begin{tabular}[t]{@{}l@{}}
NTU-VIRAL (IJRR)\\
\cite{nguyen2022ijrr-ntuviral}
\end{tabular} &
$\checkmark$ & -- & $\checkmark$ & $\checkmark$ &
Laser &
Buildings 
& Multi-sensor SLAM \\

& Aerial\&Ground &
\begin{tabular}[t]{@{}l@{}}
GRACO (RAL)\\
\cite{zhu2023ral-graco}
\end{tabular} &
$\checkmark$ & -- & $\checkmark$ & $\checkmark$ &
INS; GPS &
Campus 
& Collaborative SLAM \\

& Aerial &
\begin{tabular}[t]{@{}l@{}}
MARS-LVIG (IJRR)\\
\cite{li2024ijrr-marslvig}
\end{tabular} &
$\checkmark$ & -- & $\checkmark$ & $\checkmark$ &
RTK; DJI L1 &
Island; Town; Valley; Airfield 
& Multi-sensor SLAM \\

& Aerial &
\begin{tabular}[t]{@{}l@{}}
$D^2$SLAM (TRO)\\
\cite{xu2024tro-omni}
\end{tabular} &
-- & -- & $\checkmark$ & $\checkmark$ &
MoCap &
Office 
& Collaborative VI-SLAM \\

& Aerial &
\begin{tabular}[t]{@{}l@{}}
FIReStereo (RAL)\\
\cite{dhrafani2025ral-firestereo}
\end{tabular} &
$\checkmark$ & -- & $\checkmark$ & $\checkmark$ &
SLAM algorithm &
Woodland; Parking Lot 
& Depth; SLAM; Thermal perception \\
\midrule

\multirow{9}{*}{\rotatebox{90}{\parbox{7.2cm}{\centering\textbf{Agricultural Datasets}}}}

& Ground &
\begin{tabular}[t]{@{}l@{}}
Sugar Beets (IJRR)\\
\cite{chebrolu2017ijrr-sugarbeets}
\end{tabular} &
$\checkmark$ & -- & -- & $\checkmark$ &
Manually labelled &
Sugar-beet Fields 
& Classification; Mapping \\

& Ground &
\begin{tabular}[t]{@{}l@{}}
Rosario (IJRR)\\
\cite{pire2019ijrr-rosario}
\end{tabular} &
-- & -- & $\checkmark$ & $\checkmark$ &
RTK &
Soybean Fields 
& Multi-sensor SLAM \\

& Ground &
\begin{tabular}[t]{@{}l@{}}
Macadamia Orchard (AuRo)\\
\cite{islam2023auro-orchard}
\end{tabular} &
-- & -- & -- & $\checkmark$ &
-- &
Orchard 
& Visual SLAM  \\

& Ground &
\begin{tabular}[t]{@{}l@{}}
CitrusFarm (ISVC)\\
\cite{teng2023isvc-citrusfarm}
\end{tabular} &
$\checkmark$ & -- & $\checkmark$ & $\checkmark$ &
RTK &
Orchard 
& Crop monitoring  \\

& Ground &
\begin{tabular}[t]{@{}l@{}}
Under-Canopy (IJRR)\\
\cite{cuaran2024ijrr-undercanopy}
\end{tabular} &
-- & -- & $\checkmark$ & $\checkmark$ &
RTK & 
Corn and Soybean Fields 
& Multi-sensor SLAM  \\

& Ground &
\begin{tabular}[t]{@{}l@{}}
GREENBOT (Sensors)\\
\cite{canadas2024sensors-greenbot}
\end{tabular} &
$\checkmark$ & -- & $\checkmark$ & $\checkmark$ &
-- &
Greenhouse 
& SLAM  \\

& Ground &
\begin{tabular}[t]{@{}l@{}}
MOLO-SLAM (Agriculture)\\
\cite{lv2024agriculture-moloslam}
\end{tabular} &
$\checkmark$ & -- & $\checkmark$ & $\checkmark$ &
-- &
Grape and Tea Plantation 
& Semantic SLAM  \\

& Aerial &
\begin{tabular}[t]{@{}l@{}}
GrapeSLAM (Data Br.)\\
\cite{wang2025datainbrief-grapeslam}
\end{tabular} &
-- & -- & $\checkmark$ & $\checkmark$ &
RTK &
Vineyard 
& SLAM; SfM \\

& Aerial &
\begin{tabular}[t]{@{}l@{}}
WeedsGalore (WACV)\\
\cite{celikkan2025wacv-weedsgalore}
\end{tabular} &
-- & -- & -- & $\checkmark$ &
Manually labelled &
Cotton Fields 
& Classification; Detection; Segmentation\\
\midrule


& Aerial & 
\textbf{AgriLiRa4D} &
$\checkmark$ & $\checkmark$ & $\checkmark$ & -- &
\textbf{FINS-RTK} &
\textbf{Flat, Hilly, and Terraced Fields} 
&
\textbf{Multi-sensor SLAM} 
\\
\bottomrule
    
\end{tabularx}
}
\end{sidewaystable*}


In this section, we review the existing multi-sensor UAV-included datasets and agricultural datasets, examining their sensor configurations, collection environments, ground-truth acquisition, and typical downstream tasks, to position our proposed dataset in the broader research landscape. 
All the reviewed datasets are listed in Table~\ref{tab:datasets}.

\subsection{Multi-Sensor UAV-included Datasets}

In recent years, multi-sensor fusion has become an increasingly popular paradigm for enhancing UAV autonomous navigation and localization. 
To address the challenges posed by complex and dynamic environments, modern pipelines have evolved from early visual–inertial frameworks to comprehensive multimodal configurations that integrate RGB, depth, stereo, lidar, event cameras, and GNSS/UWB signals. 
In line with this development, the research community has introduced a number of multi-sensor UAV datasets that serve as standard benchmarks for evaluating visual–inertial, lidar–visual, and other cross-modal navigation and localization systems. 
Depending on the operating environment, existing multi-sensor UAV datasets can be broadly divided into three categories.

\textbf{Indoor or controllable environments}.
Representative datasets in this category include EuRoC~\citep{burri2016ijrr-euroc}, UPenn Fast Flight~\citep{sun2018ral-upennfastflight}, and $D^2$SLAM~\citep{xu2024tro-omni}. 
These datasets are collected in laboratories, offices, and warehouses under controlled motion profiles, stable lighting conditions, and limited environmental disturbances. 
They have become widely adopted benchmarks for evaluating highly dynamic SLAM and aggressive flight navigation.

\textbf{Urban and campus environments}.
Examples include Zurich Urban MAV~\citep{majdik2017ijrr-zurichmav}, NTU-VIRAL~\citep{nguyen2022ijrr-ntuviral}, and GRACO~\citep{zhu2023ral-graco}. 
These datasets capture flights through structured building complexes, university campuses, and road networks, offering rich 3D geometry and frequently incorporating GNSS-based ground truth. 
They serve as key benchmarks for multimodal SLAM, collaborative perception, and GNSS-assisted navigation in structured human-made environments.

\textbf{Natural or semi-natural environments}.
Representative datasets such as UZH-FPV~\citep{delmerio2019icra-uzhfpv}, MARS-LVIG~\citep{li2024ijrr-marslvig}, and FIReStereo~\citep{dhrafani2025ral-firestereo} focus on natural terrains including islands, valleys, grasslands, and forests. 
These environments introduce challenges such as drastic illumination changes, motion blur, foliage occlusion, texture sparsity, and complex, unstructured geometries, making them valuable for stress-testing perception and navigation systems in outdoor and partially unstructured scenes.

Although the aforementioned datasets have greatly advanced the development of multi-modal SLAM, they predominantly feature environments with clear man-made structures or natural geometric cues, offering stable feature points and semantic anchors for reliable perception and mapping. 
In contrast, agricultural scenes present fundamentally different and highly degraded conditions: dense and repetitive vegetation, weak textures, frequent occlusion and foliage motion, strong illumination and wind variations, and a lack of stable structural anchors. 
As a result, current datasets provide limited support for evaluating low-altitude UAV localization and mapping in real precision-agriculture environments.

\subsection{Agricultural Datasets}

Agricultural datasets have historically emerged from the needs of crop growth monitoring and precision farming, where the primary objectives include plant phenotyping, vegetation segmentation, disease detection, and field-level condition assessment. 
Only in recent years has the community begun to introduce datasets specifically designed for agricultural SLAM and autonomous navigation. 
Compared with generic robotics datasets, agricultural datasets are uniquely shaped by crop structure, seasonal variations, and operational constraints, leading to diverse sensing setups and scene characteristics. 
Existing agricultural datasets can be grouped into four major categories according to platform type, scene structure, and intended task.

\textbf{Phenotyping and crop-monitoring datasets}.
Early agricultural datasets, such as Sugar Beets~\citep{chebrolu2017ijrr-sugarbeets}, focus on multi-spectral, RGB-D, and lidar measurements for crop classification, inter-row localization, and biomass estimation. 
These datasets provide rich multimodal signals for static analysis and plant-level tasks, but they are not designed for evaluating UAV-based SLAM pipelines.

\textbf{Ground robotic SLAM datasets in open farmland}.
Datasets including Rosario~\citep{pire2019ijrr-rosario}, Under-Canopy~\citep{cuaran2024ijrr-undercanopy}, and GREENBOT~\citep{canadas2024sensors-greenbot} capture near-ground navigation in soybean fields, corn rows, and greenhouses using stereo cameras, wheel odometry, IMU, lidar, and GNSS. 
These datasets rely heavily on structured row geometry and near-field texture cues, enabling reliable perception under constrained viewpoints but offering limited diversity in crop types and environmental conditions.

\textbf{Ground datasets in orchards and plantations}.
Datasets such as Macadamia Orchard~\citep{islam2023auro-orchard}, CitrusFarm~\citep{teng2023isvc-citrusfarm}, and MOLO-SLAM~\citep{lv2024agriculture-moloslam} are collected in orchards and vineyards, where trees or vines form clear vertical structures and repeated spatial patterns. 
These environments introduce moderate natural variability while still providing stable geometric anchors (tree trunks, trellis systems), which simplify localization compared with open-field row crops.

\textbf{Limited UAV-based agricultural datasets}.
A small number of agricultural UAV datasets, such as GrapeSLAM~\citep{wang2025datainbrief-grapeslam}, explore aerial views over vineyards but remain limited in crop type, spatial scale, and overall scene diversity. 
The WeedsGalore dataset~\citep{celikkan2025wacv-weedsgalore} broadens sensing with multispectral and multi-temporal imagery, yet the absence of lidar, full 6-DoF ground truth, and richer geometric structure still restricts its suitability for benchmarking SLAM or reconstruction algorithms in more demanding agricultural settings.

Despite their contributions, the aforementioned datasets predominantly focus on ground platforms, constrained viewpoints, and structured or semi-structured agricultural environments. 
Most datasets involve a single crop species, a single season, or a limited geographic region, with few providing long-term, multi-season, or multi-terrain coverage. 
Furthermore, existing UAV-based agricultural datasets remain scarce and are mostly collected in orchards, where stable vertical structures simplify perception. 
As a result, they fall short of capturing the dense foliage, highly repetitive textures, foliage-induced motion, strong illumination fluctuations, and other degraded conditions commonly encountered by UAVs operating at low altitude over open-field crops.


\section{System Overview} \label{sec:system}
This section summarizes the AgriLiRa4D system configuration, including the onboard sensor setup, multi-sensor calibration, and released data formats. 
Together, these components define the sensing architecture, inter-sensor geometric relationships, and data organization of the dataset.

\subsection{Sensor Setup}

\begin{figure*}[htbp]
    \centering
    \begin{subfigure}[t]{0.46\textwidth}
        \centering
        \includegraphics[width=\linewidth]{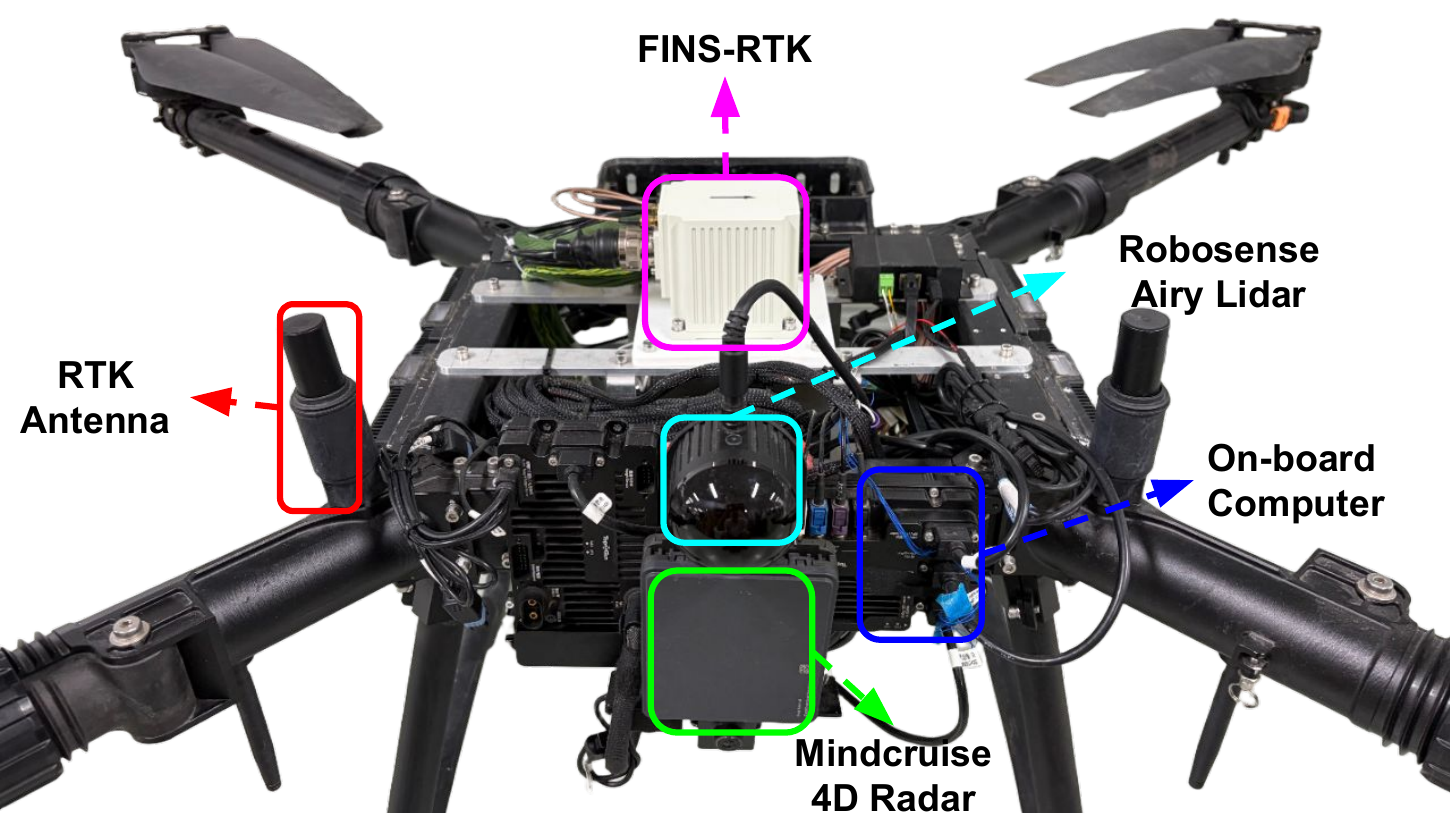}
        \caption{Sensor setup on the TopXGun FP300E platform.}
        \label{fig:fp300e_sensor_setup}
    \end{subfigure}
    \hfill
    \begin{subfigure}[t]{0.52\textwidth}
        \centering
        \includegraphics[width=\linewidth]{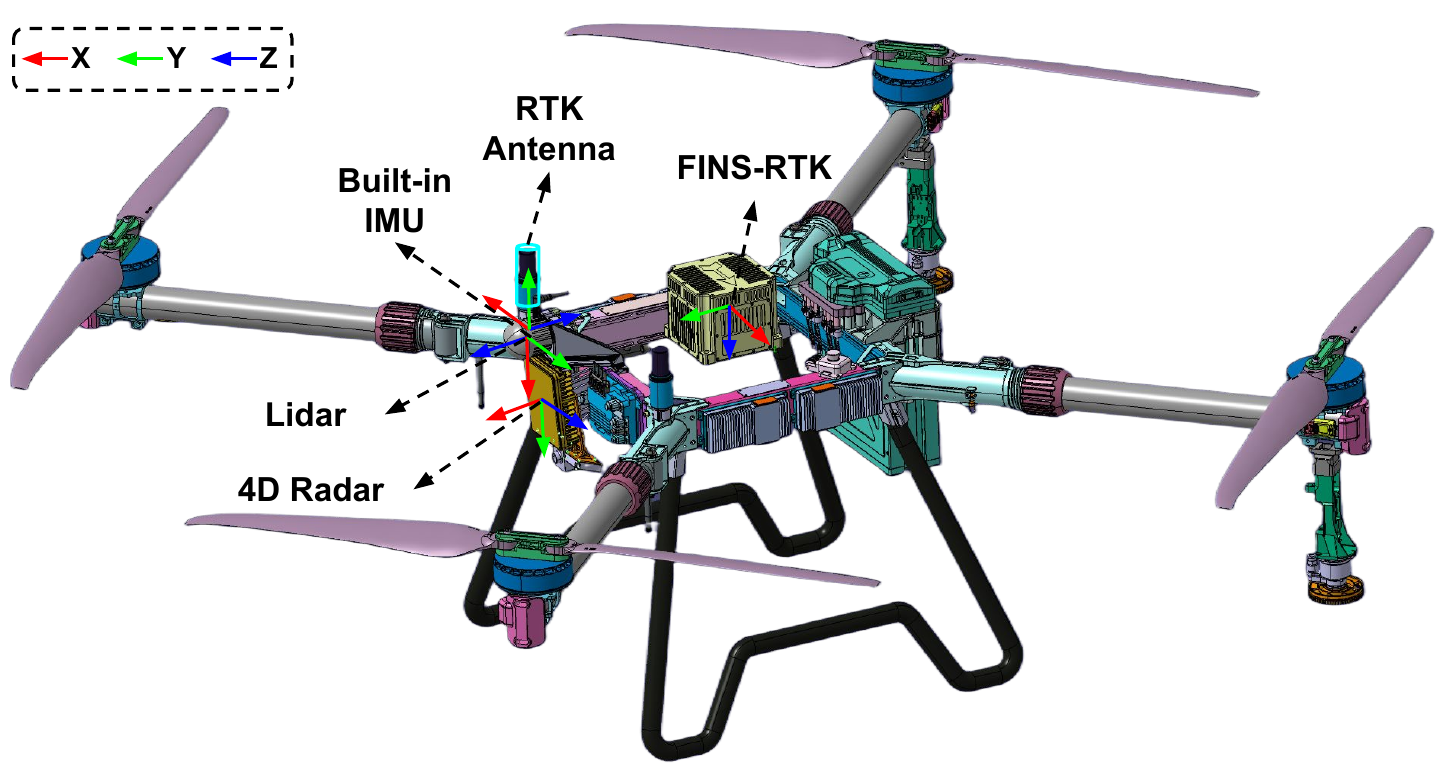}
        \caption{Relative positions and coordinate frames of all sensors.}
        \label{fig:fp300e_sensor_tf}
    \end{subfigure}

    \caption{\textbf{Sensor configuration on the TopXGun FP300E.}
    The left panel shows the real UAV platform used for data collection, while the right panel illustrates the relative positions and coordinate frames of all sensors.}
    \label{fig:fp300e_overview}
\end{figure*}

A TopXGun FP300E agricultural UAV~\citep{txgfp300e} carries a customized SLAM payload consisting of a 3D lidar (with an integrated IMU) and a 4D radar. 
The platform is equipped with a FINS-RTK navigation module, providing centimeter-level position accuracy and high-fidelity orientation ground truth.
All sensor data are logged by an on-board ARM computer based on the RK3588 processor and running the Robot Operating System (ROS). 
The sensors interface with the computer over Gigabit Ethernet and are synchronized using the IEEE 1588 Precision Time Protocol (PTP). 
The onboard RK3588 serves as the PTP master, with the lidar, 4D radar, and FINS-RTK operating as slaves.
The overall hardware configuration is shown in Figure~\ref{fig:fp300e_sensor_setup}, and the specifications of each sensor are summarized below.

\begin{figure}[t]
    \centering
    \begin{subfigure}[t]{0.48\linewidth}
        \centering
        \includegraphics[width=\linewidth]{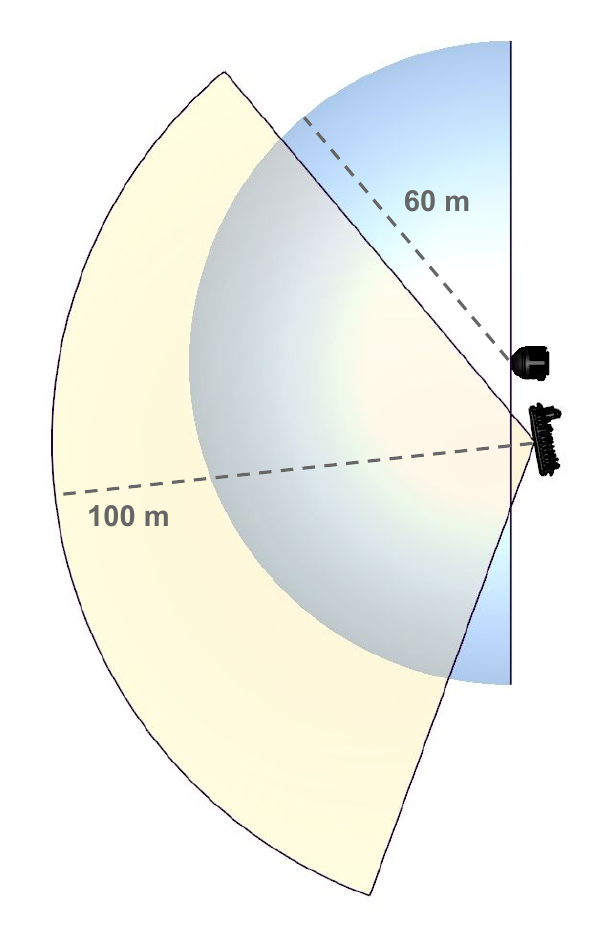}
        \caption{Side view}
        \label{fig:fov_side-view}
    \end{subfigure}
    \hfill
    \begin{subfigure}[t]{0.48\linewidth}
        \centering
        \includegraphics[width=\linewidth]{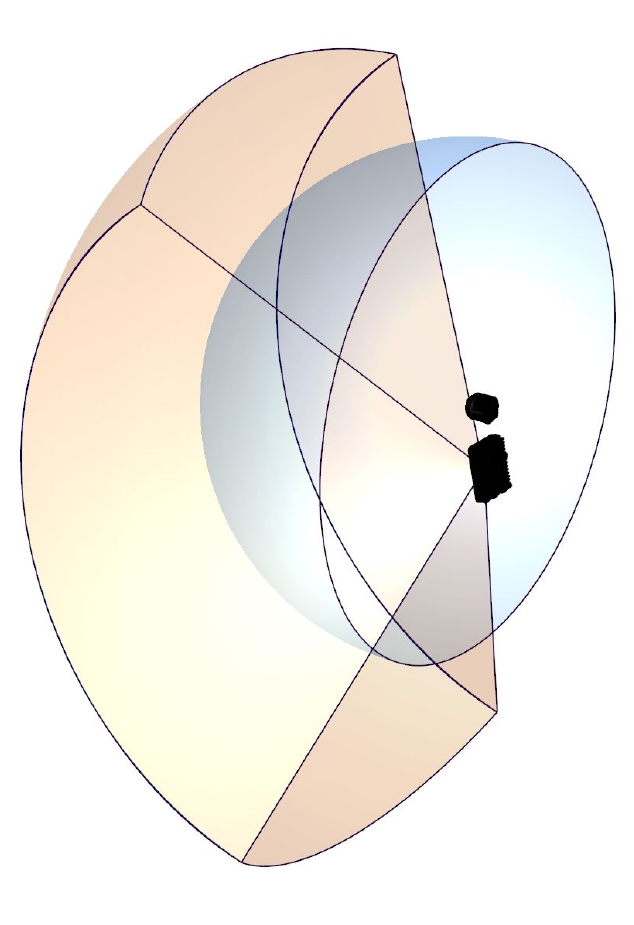}
        \caption{Oblique view}
        \label{fig:fov_oblique-view}
    \end{subfigure}
    \caption{\textbf{Visualization of the FoV configuration for the lidar and 4D radar.} 
    The two viewpoints illustrate their respective sensing coverages, with the lidar rendered in \textcolor{blue}{blue} and the 4D radar in \textcolor{yellow}{yellow}.
    The FoV extents approximately correspond to maximum sensing ranges of 60~m for the lidar and 100~m for the 4D radar.}
    \label{fig:sensor_fov}
\end{figure}

\begin{enumerate}
    \item \textbf{3D Lidar.} 
    We employ the RoboSense Airy~\citep{airy} as a lightweight and cost-effective 3D lidar. 
    As shown in Figure~\ref{fig:sensor_fov}, it provides a $96$-beam configuration with a $90\unit{\degree}\times 360\unit{\degree}$ FoV and a maximum range of \qty{60}{\meter} (or \qty{30}{\meter} at \qty{10}{\percent} reflectivity in outdoor conditions). 
    The sensor operates at \qty{10}{\hertz} and outputs point clouds with timestamps, ring indices, and reflectivity values. 
    Its large vertical FoV is a key reason for selection, as narrower-coverage alternatives such as Velodyne VLP-16 (\qty{30}{\degree}) or Livox Mid-360 (\qty{59}{\degree}) may fail to capture sufficient structure in cluttered agricultural environments. 
    
    \item \textbf{IMU.}
    The lidar integrates an IIM-42652 IMU~\citep{IIM42652}, providing angular velocity and linear acceleration at \qty{200}{\hertz}. 
    The gyroscope exhibits a bias instability below \qty{3.6}{\degree\per\hour} and a temperature coefficient of offset (TCO) under \qty{0.02}{dps\per\kelvin}. 
    The accelerometer offers a TCO of \qty{0.15}{\milli\gram\per\kelvin} and low spectral noise of \qty{70}{\micro\gram\per\sqrt{\hertz}}.
    IMU measurements are transmitted via the same Ethernet interface as the lidar and logged synchronously in ROS.
    
    \item \textbf{4D Radar.} 
    The Mindcruise A1 4D radar is mounted directly beneath the lidar to maximize FoV overlap and improve cross-sensor observability, as shown in Figure~\ref{fig:fp300e_sensor_setup} and~\ref{fig:sensor_fov}. 
    It offers a $60\unit{\degree}\times 120\unit{\degree}$ FoV and a detection range of up to \qty{100}{\meter}. 
    In addition to range and angle, the sensor provides doppler velocity measurements from \qty{-35}{\meter\per\second} to \qty{20}{\meter\per\second}, enabling the direct observation of dynamic targets. 
    Operating at \qty{10}{\hertz}, the radar delivers reliable performance in dusty or foggy environments due to the longer wavelength of millimeter-wave signals, complementing the lidar in challenging agricultural scenarios.
    
    \item \textbf{FINS-RTK.} 
    High-precision ground-truth poses are provided by the TJ-FINS70D FINS-RTK module, which fuses FINS inertial measurements with a built-in RTK receiver. 
    The system achieves position accuracy better than \qty{2}{\centi\meter} $+$ \qty{1}{ppm} (50\% CEP), and orientation accuracy of \qty{0.01}{\degree} (\qty{1}{\sigma}) in roll/pitch and \qty{0.05}{\degree} (\qty{1}{\sigma}) in yaw. 
    Operating at \qty{100}{\hertz}, this module supplies a reliable reference for quantitative evaluation of SLAM and odometry performance.

\end{enumerate}

\subsection{Sensor Calibration}

\begin{figure}[htbp]
    \centering

    \begin{subfigure}[t]{0.85\linewidth}
        \centering
        \includegraphics[width=\linewidth]{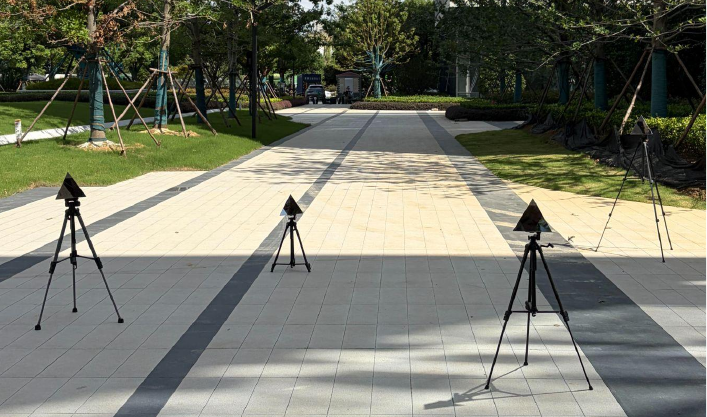}
        \caption{Corner-reflector target setup.}
        \label{fig:calib_target_setup}
    \end{subfigure}

    \begin{subfigure}[t]{0.85\linewidth}
        \centering
        \includegraphics[width=\linewidth]{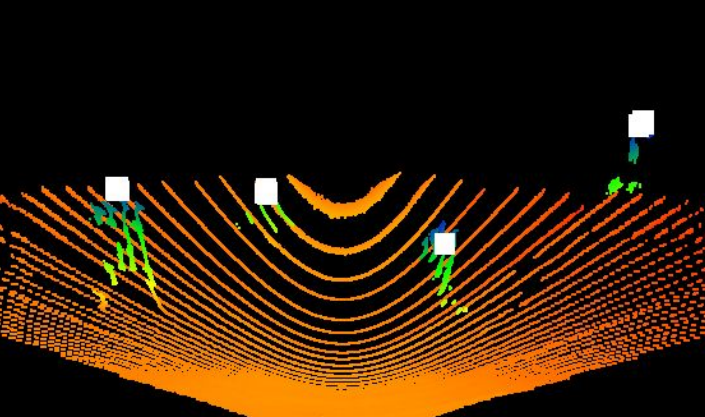}
        \caption{Lidar-radar alignment after refinement.}
        \label{fig:calib_alignment}
    \end{subfigure}

    \caption{
    \textbf{Lidar-radar extrinsic calibration setup and result.}
    (a) Four corner reflectors placed at different ranges and heights within the overlapping sensing region.
    (b) Qualitative visualization of lidar-radar registration around the calibration targets, with height-colored lidar points and white radar points.}
    \label{fig:extrinsic_calibration}
\end{figure}

\begin{figure*}[t]
    \centering
    \setlength{\tabcolsep}{2pt}
    \renewcommand{\arraystretch}{1.0}

    \begin{tabular}{c c c c}
        & \textbf{Scene} & \textbf{Side View} & \textbf{Top-down View} \\
        \rotatebox{90}{\parbox{4.2cm}{\centering\textbf{Ground}}} &
        \includegraphics[width=0.32\textwidth,height=0.24\textwidth]{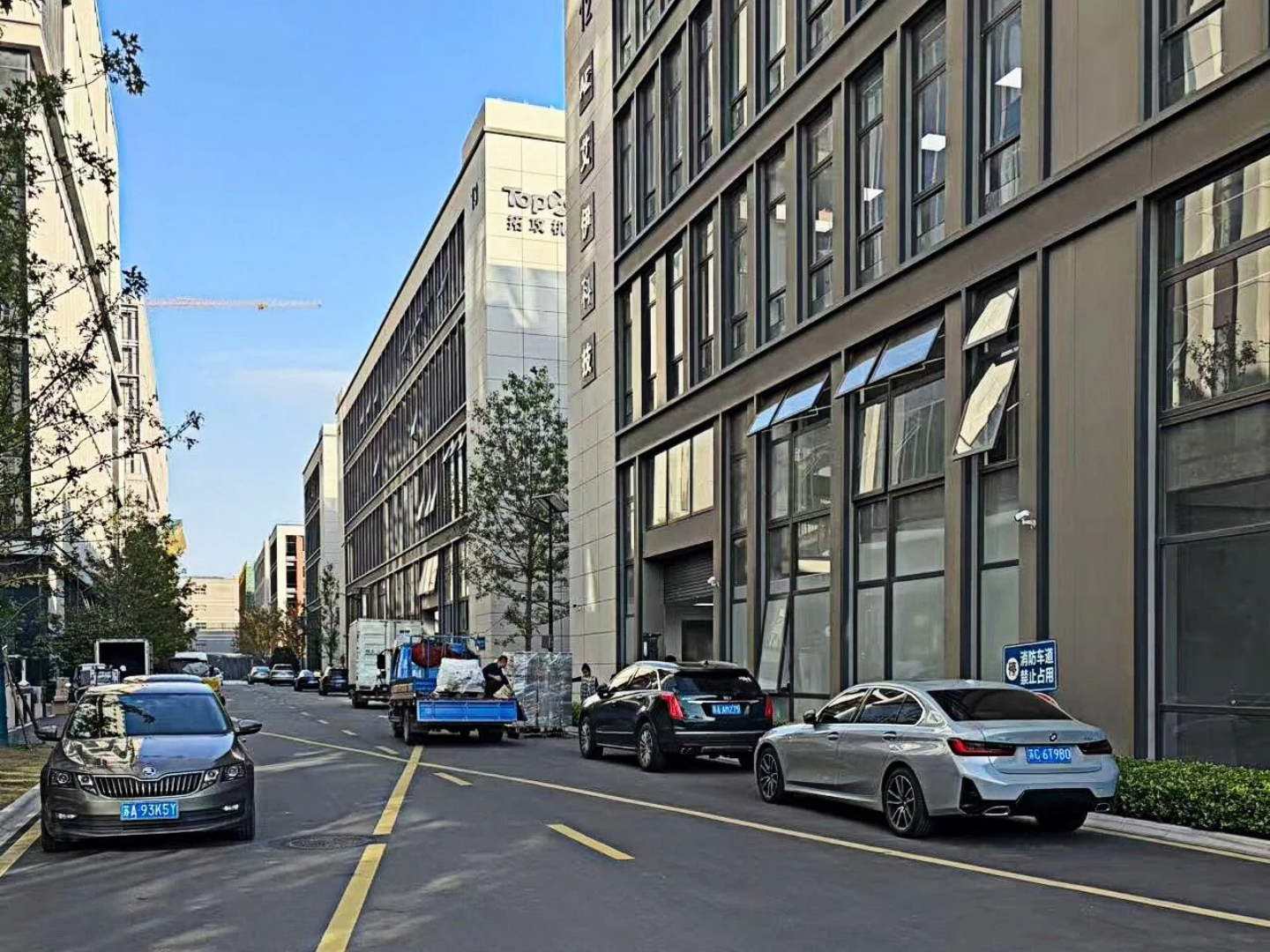} &
        \includegraphics[width=0.32\textwidth,height=0.24\textwidth]{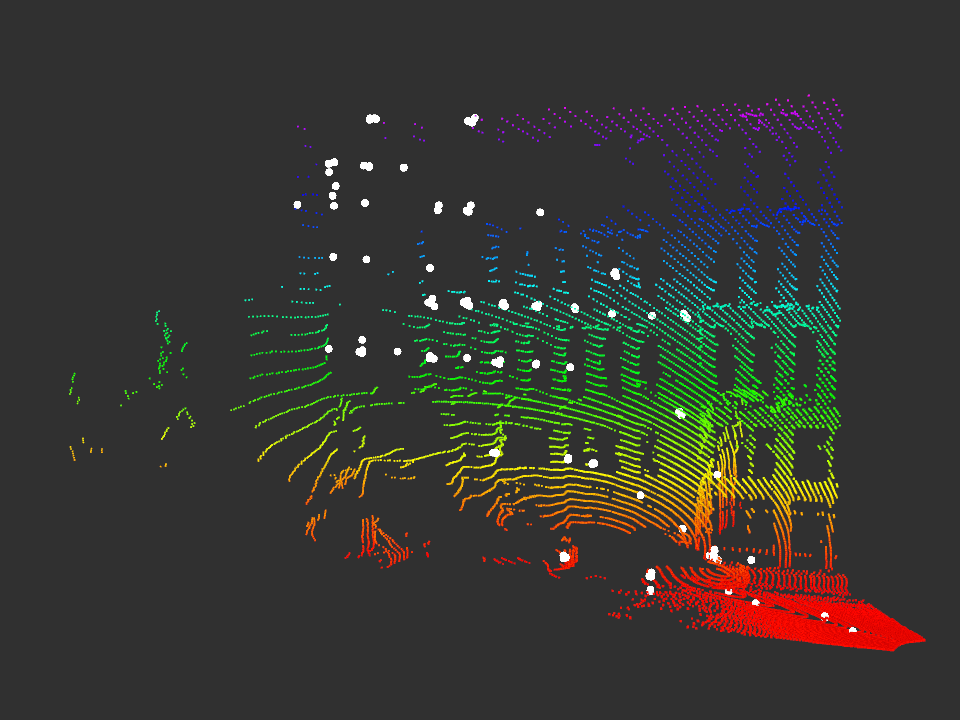} &
        \includegraphics[width=0.32\textwidth,height=0.24\textwidth]{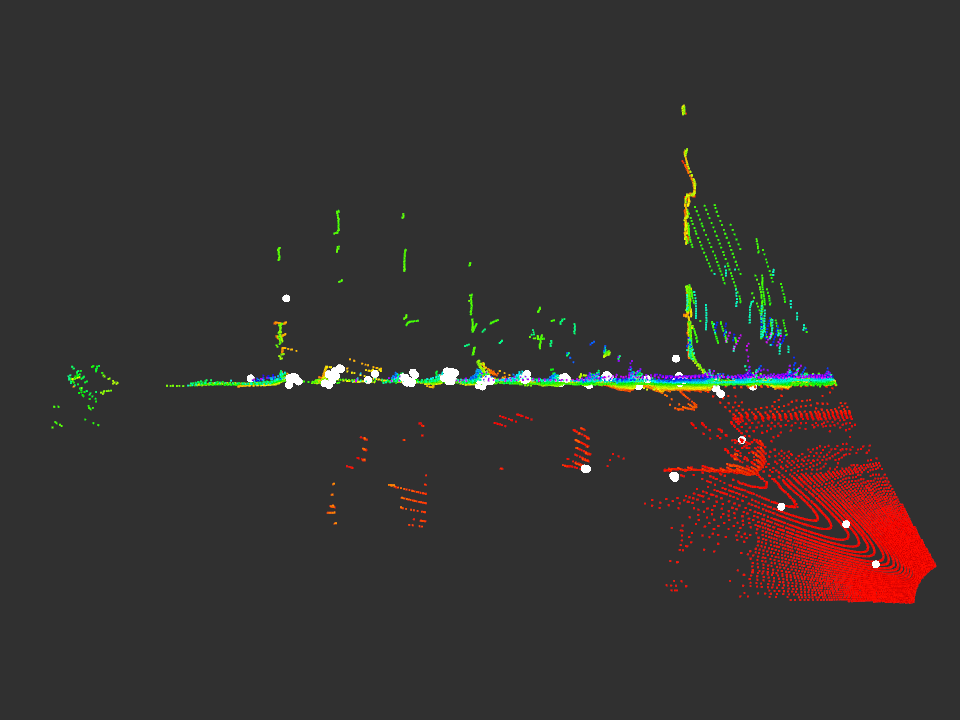} \\
        \rotatebox{90}{\parbox{4.2cm}{\centering\textbf{Hovering}}} &
        \includegraphics[width=0.32\textwidth,height=0.24\textwidth]{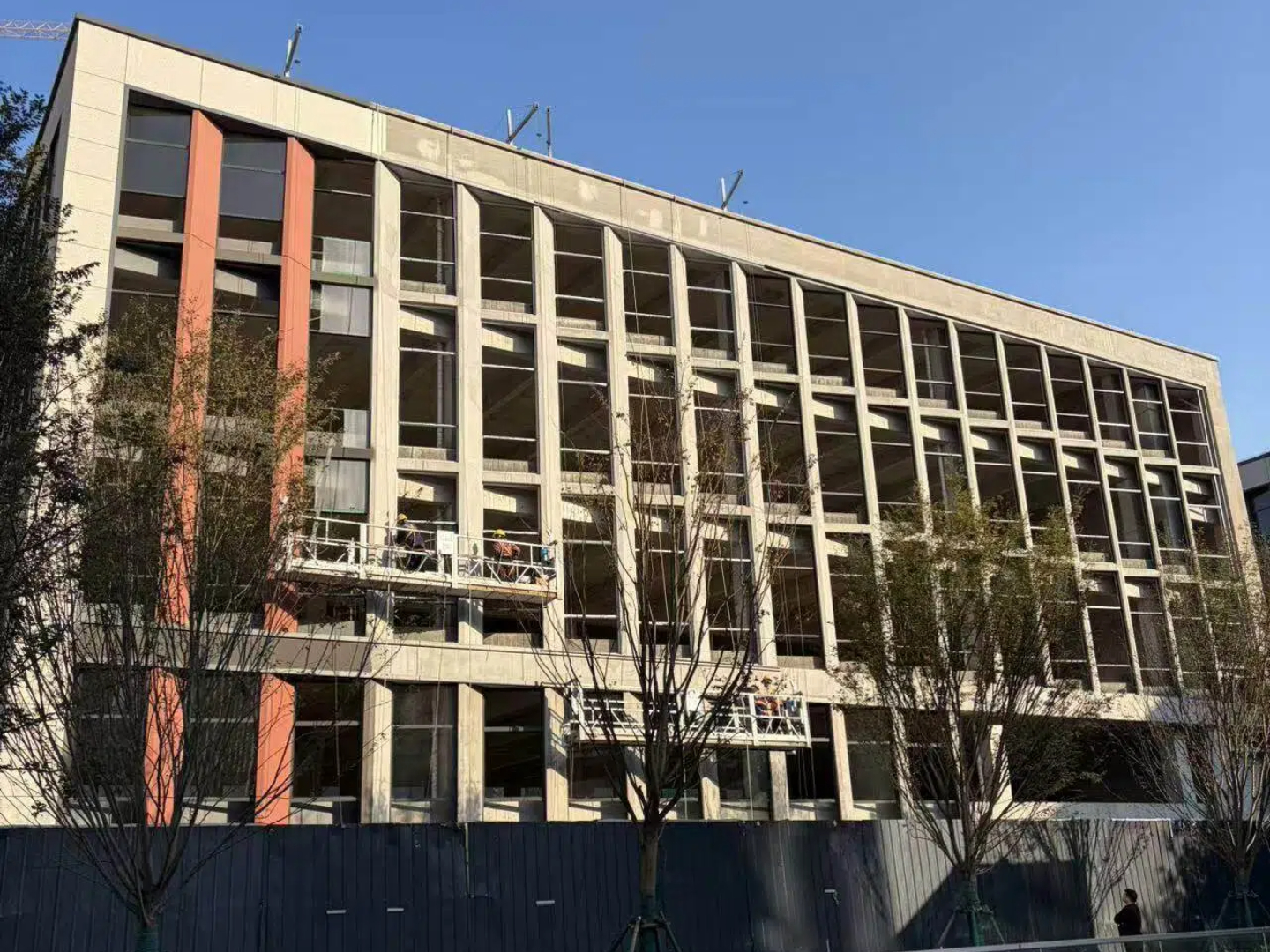} &
        \includegraphics[width=0.32\textwidth,height=0.24\textwidth]{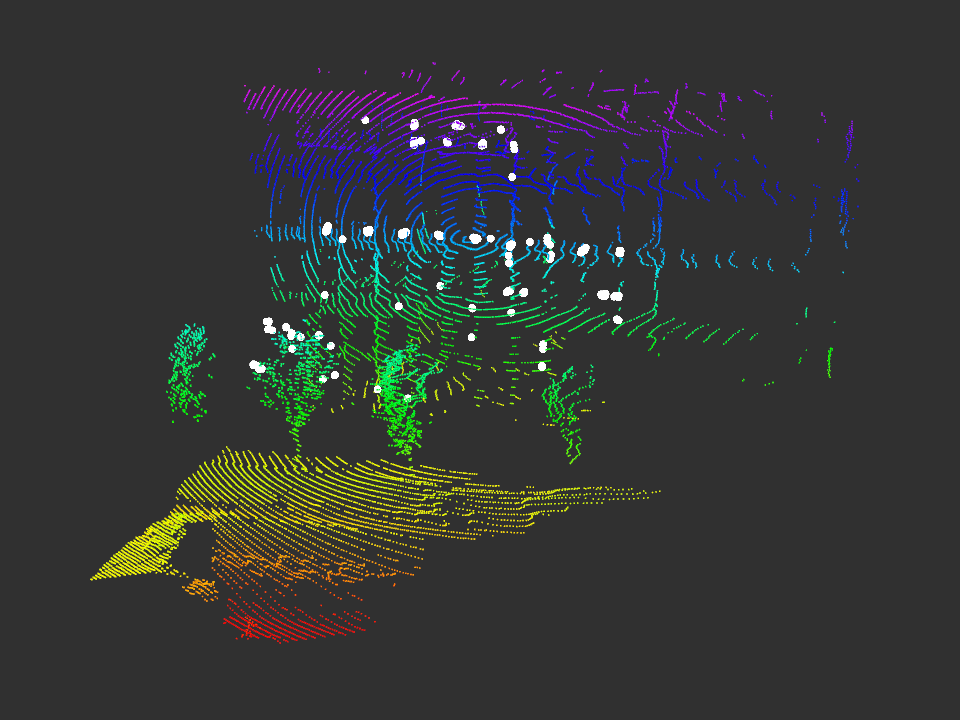} &
        \includegraphics[width=0.32\textwidth,height=0.24\textwidth]{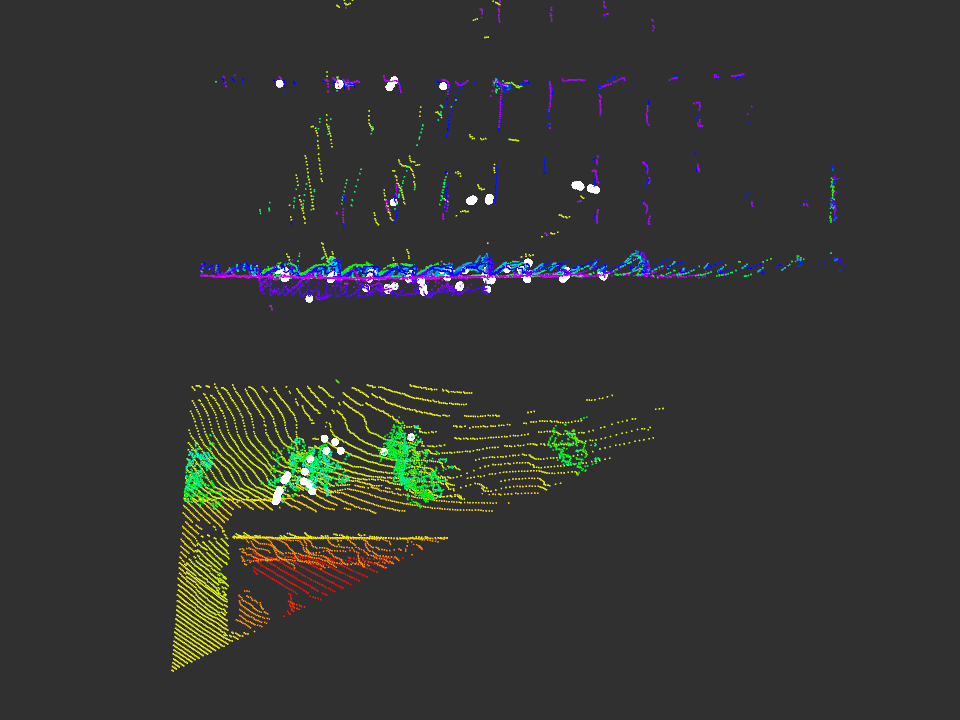} \\
    \end{tabular}
    \caption{\textbf{Visualization of the lidar and 4D radar point clouds used to assess the extrinsic calibration.} 
    Height-colored lidar points and white 4D radar points are visualized in a common frame following extrinsic alignment, with side and top-down views illustrating the spatial consistency across scenarios.}
    \label{fig:aligned_pointclouds}
\end{figure*}

The extrinsic parameters among the lidar, the 4D radar, the RTK antenna phase centers, and the FINS-RTK module are derived directly from the UAV's CAD design files and released together with the dataset (see Figure~\ref{fig:fp300e_sensor_tf}). 
The corresponding lever-arm offsets between the RTK antenna phase centers and the FINS-RTK module are configured in the FINS-RTK system prior to data collection and internally compensated during GNSS/INS fusion.
Because the internal IMU is rigidly integrated within the lidar unit, its relative transform is factory-calibrated and remains constant during data collection.

Accurate lidar-radar extrinsic calibration is crucial for multi-sensor fusion and consistent cross-modal point cloud alignment. We formulate this calibration as a 3D–3D registration problem, aligning the coordinate frames of both sensors within a unified reference. 
The CAD-derived translation and rotation serve as initial priors, which are subsequently refined through a manual calibration procedure~\citep{opencalib} using four corner reflectors placed at different ranges and heights within the overlapping sensing region of the lidar and 4D radar. 
Observations are collected from five sensor poses, yielding a total of 20 lidar-radar target correspondences for quantitative calibration evaluation.
The corner reflectors are identified independently in the lidar and radar point clouds, and correspondences are established according to the known target arrangement.

For each correspondence, the target-registration residual is computed as the Euclidean distance between the lidar-detected reflector position and the corresponding radar detection transformed into the lidar frame. 
The registration RMSEs obtained using the CAD-derived and refined extrinsics are compared across the five sensor poses in Table~\ref{tab:extrinsic_residual}. 
As no ground-truth extrinsic transformation is available, the reported residuals characterize the relative improvement in lidar-radar target alignment rather than the absolute extrinsic error, and may also reflect sensor measurement and reflector-localization uncertainties.
Figure~\ref{fig:extrinsic_calibration} shows the corner-reflector arrangement and a qualitative visualization of the lidar-radar registration around the calibration targets.

\begin{table}[t]
    \centering
    \caption{\textbf{Lidar-radar target registration residuals before and after extrinsic refinement.}
    RMSE, Mean, and Max. are computed from the 3D Euclidean registration residuals. All values are reported in meters.}
    \label{tab:extrinsic_residual}

    \small
    \setlength{\tabcolsep}{5pt}
    \renewcommand{\arraystretch}{1.1}

    \begin{tabular}{lcccccc}
        \toprule
        \multirow{2}{*}{\textbf{Trial}} &
        \multicolumn{3}{c}{\textbf{CAD Prior}} &
        \multicolumn{3}{c}{\textbf{Refined}} \\
        \cmidrule(lr){2-4}
        \cmidrule(lr){5-7}
        &
        \textbf{RMSE} &
        \textbf{Mean} &
        \textbf{Max.} &
        \textbf{RMSE} &
        \textbf{Mean} &
        \textbf{Max.} \\
        \midrule

        Trial 1 & 0.157 & 0.144 & 0.197
                & 0.084 & 0.075 & 0.130 \\

        Trial 2 & 0.176 & 0.174 & 0.208
                & 0.090 & 0.086 & 0.126 \\

        Trial 3 & 0.147 & 0.145 & 0.181 
                & 0.101 & 0.095 & 0.134 \\

        Trial 4 & 0.151 & 0.145 & 0.178 
                & 0.100 & 0.082 & 0.164 \\

        Trial 5 & 0.135 & 0.132 & 0.164 
                & 0.094 & 0.078 & 0.129 \\

        \midrule
        \textbf{Overall} &  \textbf{0.154} & \textbf{0.148} & \textbf{0.208}
                        & \textbf{0.094} & \textbf{0.083} & \textbf{0.164} \\

        \bottomrule
    \end{tabular}
\end{table}

Once calibration is completed, lidar and 4D radar point clouds are transformed into the shared coordinate frame. 
Figure~\ref{fig:aligned_pointclouds} further illustrates the cross-modal alignment in a static ground sequence and a UAV hovering sequence, with only the overlapping sensing regions displayed for clarity. 
These real-scene examples complement the quantitative target-based calibration evaluation above.

\begin{figure}[htbp]
    \centering
    \includegraphics[width=0.48\textwidth]{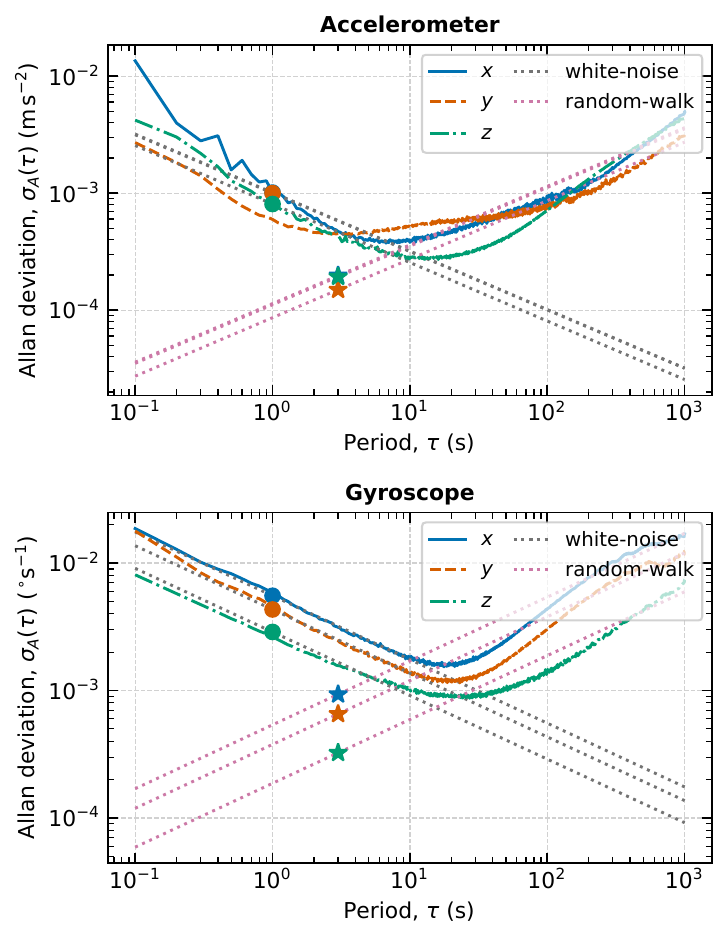}

    \caption{\textbf{Allan deviation analysis of the integrated IMU.}
    The accelerometer and gyroscope curves are used to estimate the corresponding white-noise and random-walk parameters.}
    \label{fig:allan_variance}
\end{figure}

\begin{table}[t]
    \centering
    \caption{\textbf{Estimated stochastic noise parameters of the integrated IMU based on Allan variance analysis.}}
    \label{tab:imu_calibration}

    \small
    \setlength{\tabcolsep}{8pt}
    \renewcommand{\arraystretch}{1.1}

    \begin{tabular}{l c c}
        \toprule
        \textbf{Parameter} & \textbf{Value} & \textbf{Unit} \\
        \midrule
        
        Accel. White Noise
        & $1.0110 \times 10^{-3}$
        & $\mathrm{m/s^2/\sqrt{Hz}}$ \\

        Accel. Random Walk
        & $1.9743 \times 10^{-4}$
        & $\mathrm{m/s^3/\sqrt{Hz}}$ \\

        Gyro. White Noise
        & $9.6675 \times 10^{-5}$
        & $\mathrm{rad/s/\sqrt{Hz}}$ \\

        Gyro. Random Walk
        & $1.6278 \times 10^{-5}$
        & $\mathrm{rad/s^2/\sqrt{Hz}}$ \\
        
        \bottomrule
    \end{tabular}
\end{table}

To further characterize the stochastic noise of the integrated IMU, we perform Allan variance analysis~\citep{AllanVarianceRos} using more than three hours of stationary measurements collected under stable operating conditions.
Figure~\ref{fig:allan_variance} shows the Allan deviation curves and corresponding noise-model fits for both the accelerometer and gyroscope, from which the white-noise and random-walk components are estimated using the -1/2 and 1/2 slope regions, respectively.
The resulting gyroscope and accelerometer white-noise densities and random-walk noise densities are reported in Table~\ref{tab:imu_calibration} and released with the dataset to support reproducible configuration of inertial-aided SLAM algorithms.

\subsection{Temporal Synchronization}

All onboard sensors are synchronized through IEEE 1588 PTP, with the RK3588 computer serving as the master clock and the lidar, 4D radar, and FINS-RTK operating as slaves. 
To quantitatively evaluate the temporal synchronization performance, we measure the end-to-end timestamp synchronization residuals over a continuous one-hour period using the PTP message timestamps and the corresponding packet-capture timestamps at the RK3588 network interface.

Specifically, let $t_1$ denote the precise timestamp extracted from a PTP Follow\_Up message and $T_1$ the packet-capture timestamp of the corresponding Sync message. 
For a sensor data packet, $t_2$ denotes its embedded sensor timestamp and $T_2$ the corresponding packet-capture timestamp. 
The temporal synchronization residual is defined as
\begin{equation}
    \Delta t = (t_2-t_1)-(T_2-T_1).
    \label{eq:sync_residual}
\end{equation}
The resulting mean, standard deviation, and maximum absolute residuals for the lidar, 4D radar, and FINS-RTK are summarized in Table~\ref{tab:time_sync}.

\begin{table}[t]
    \centering
    \caption{\textbf{End-to-end temporal synchronization residuals of the onboard sensors.}
    All values are reported in milliseconds.}
    \label{tab:time_sync}

    \small
    \setlength{\tabcolsep}{10pt}
    \renewcommand{\arraystretch}{1.1}
    
    \begin{tabular}{lccc}
        \toprule
        \textbf{Sensor} &
        \textbf{Mean} &
        \textbf{Std.} &
        \textbf{Max. Abs.} \\
        \midrule
        Lidar     & 0.765 & 0.045 & 0.892 \\
        4D Radar  & 0.812 & 0.061 & 0.953 \\
        FINS-RTK  & 0.747 & 0.043 & 0.972 \\
        \bottomrule
    \end{tabular}
\end{table}

\subsection{Data Format}

The original multi-sensor measurements are provided as ROS bag files to preserve the synchronized ROS-based data streams and facilitate direct playback within standard robotics workflows. 
In addition, AgriLiRa4D provides a ROS-independent release in human-readable formats.
The corresponding ROS topics, message types, update rates, and released data formats are summarized in Table~\ref{tab:ros_topics}.

The lidar provides 3D point clouds containing $(x, y, z)$ coordinates with per-point attributes including intensity, ring index, and precise timestamps. 
The IMU supplies raw inertial measurements consisting of linear acceleration and angular velocity, where the released acceleration values are expressed in units of $g$ ($1g = 9.80665~\mathrm{m/s^2}$).
The 4D radar generates point clouds enriched with range–angle information, doppler velocity, signal-to-noise ratio (SNR), and radar cross section (RCS) values, enabling both geometric and motion-related perception.

Original timestamps are retained in the ROS-independent release to preserve temporal consistency across modalities. 
Conversion utilities, ROS-independent parsers, and detailed format documentation are provided in the public repository to facilitate direct access and reproducible use.

\begin{figure}[htbp]
    \centering
    \includegraphics[width=0.5\textwidth]{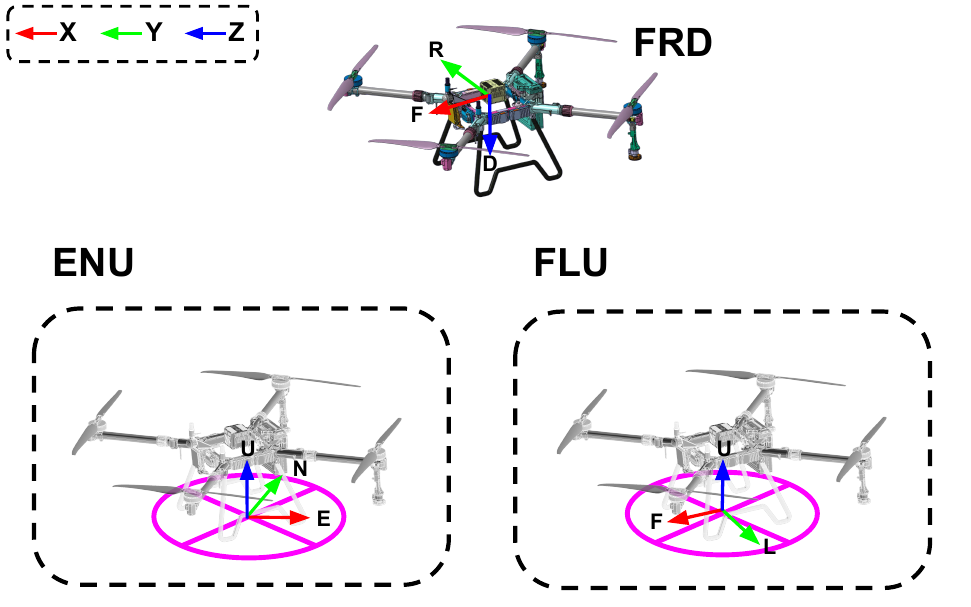}
    \caption{\textbf{Ground-truth reference frames used in this work.}
    The UAV body frame (FRD) and the two frames relative to the take-off point, FLU and ENU, are shown for defining consistent trajectory coordinates.}
    \label{fig:gt_pose}
\end{figure}

To obtain high-precision and easily usable ground-truth trajectories, 
the original FINS-RTK geodetic positions are first corrected to the integrated IMU origin by compensating for the physical lever arm using the FINS-RTK orientation and the corresponding extrinsic translation. 
The resulting IMU-referenced ground truth is subsequently provided in three forms:
(1) a pose reference expressed in the East–North–Up (ENU) frame relative to the take-off location;
(2) a pose reference expressed in the Forward–Left–Up (FLU) body frame, also anchored at the take-off point; and
(3) a geodetic reference containing latitude, longitude, and altitude in the global coordinate system.

Orientation measurements follow the Front–Right–Down (FRD) aerospace convention. 
The relationships among these coordinate frames are visualized in Figure~\ref{fig:gt_pose}. 
These unified ground-truth representations provide a consistent spatial reference for benchmarking multi-sensor SLAM algorithms.
Throughout all released sequences, the FINS-RTK system maintained a fixed RTK solution using network RTK corrections, with no GNSS outages observed during data collection. 
The released ground-truth trajectories are derived directly from the FINS-RTK outputs, with no post-processing other than the lever-arm compensation and coordinate-frame transformations described above. 
No trajectory segments were removed or flagged for degraded ground-truth quality, and additional covariance or solution-quality indicators were not retained because the RTK solution remained fixed throughout the valid sequences.

\begin{table*}[t]
    \centering
    \caption{
    \textbf{Overview of the sensors and released data formats in AgriLiRa4D.}
    Each sensor is presented with its associated ROS topic, message type, update rate, and ROS-independent data format.
    }
    
    \small
    \setlength{\tabcolsep}{5pt}
    \renewcommand{\arraystretch}{1.1}
    
    \begin{tabular}{l l l l c l}
    \toprule
    \textbf{Sensor} & \textbf{Module} & \textbf{Topic Name} & \textbf{Message Type} & \textbf{Rate (Hz)} & \textbf{Format}\\
    \midrule
    Lidar & Robosense Airy & \texttt{/rslidar\_points} & \texttt{sensor\_msgs/PointCloud2} & 10 & ASCII PCD\\
    IMU & Built-in (Lidar) & \texttt{/rslidar\_imu\_data} & \texttt{sensor\_msgs/Imu} & 200 & CSV\\
    4D Radar & Mindcruise A1 & \texttt{/radar\_points} & \texttt{sensor\_msgs/PointCloud2} & 10 & ASCII PCD\\
    \multirow[t]{3}{*}{FINS-RTK} & \multirow[t]{3}{*}{TJ-FINS70D} & \texttt{/aircraft\_pose\_enu} & \texttt{geometry\_msgs/PoseStamped} & 100 & CSV\\
     & & \texttt{/aircraft\_pose\_flu} & \texttt{geometry\_msgs/PoseStamped} & 100 & CSV\\
     & & \texttt{/aircraft\_position\_llh} & \texttt{sensor\_msgs/NavSatFix} & 100 & CSV\\
    \bottomrule
    \end{tabular}
    \label{tab:ros_topics}
\end{table*}


\section{Dataset Characteristics} \label{sec:dataset}

\begin{table*}[t]
    \centering
    \caption{\textbf{Detailed flight configurations and vegetation characteristics of all sequences.}
    The sequences are grouped according to terrain category and scanning mode.
    The reported altitude and speed values correspond to predefined flight targets.}
    
    \small
    \setlength{\tabcolsep}{5pt}
    \renewcommand{\arraystretch}{1.1}
    
    \begin{tabular}{l l l l c c c}
        \toprule
        \textbf{Scene} &
        \textbf{Vegetation} &
        \textbf{Sequence} & 
        \textbf{Scanning Mode} & 
        \textbf{Altitude (m)} & 
        \textbf{Speed (m/s)} & 
        \textbf{Path Length (m)} \\
        \midrule
        
        \multirow[c]{10}{*}{Flat Farmland} & \multirow[l]{10}{*}{Dense tall sorghum} & \texttt{NJFlatB01} & \texttt{boundary} & 5 & 3 & 434.77 \\
        & & \texttt{NJFlatB02} & \texttt{boundary} & 5 & 8 & 464.21 \\
        & & \texttt{NJFlatB03} & \texttt{boundary} & 10 & 3 & 456.32 \\
        & & \texttt{NJFlatB04} & \texttt{boundary} & 10 & 8 & 462.18 \\
        & & \texttt{NJFlatB05} & \texttt{boundary} & 15 & 3 & 465.89 \\
        & & \texttt{NJFlatB06} & \texttt{boundary} & 15 & 8 & 454.21 \\
        \cmidrule(lr){3-7}
        & & \texttt{NJFlatC01} & \texttt{coverage} & 5 & 8 & 805.65 \\
        & & \texttt{NJFlatC02} & \texttt{coverage} & 10 & 3 & 801.17 \\
        & & \texttt{NJFlatC03} & \texttt{coverage} & 10 & 8 & 798.96 \\
        & & \texttt{NJFlatC04} & \texttt{coverage} & 15 & 3 & 822.23 \\
        \midrule    
        
        \multirow[c]{12}{*}{Hilly Farmland} & \multirow[l]{12}{*}{\shortstack[l]{Lower-slope sorghum;\\upper-slope grass}} & \texttt{NJHillB01} & \texttt{boundary} & 8 & 3 & 490.61 \\
        & & \texttt{NJHillB02} & \texttt{boundary} & 8 & 8 & 493.07 \\
        & & \texttt{NJHillB03} & \texttt{boundary} & 13 & 3 & 480.98 \\
        & & \texttt{NJHillB04} & \texttt{boundary} & 13 & 8 & 484.60 \\
        & & \texttt{NJHillB05} & \texttt{boundary} & 18 & 3 & 483.84 \\
        & & \texttt{NJHillB06} & \texttt{boundary} & 18 & 8 & 488.41 \\
        \cmidrule(lr){3-7}
        & & \texttt{NJHillC01} & \texttt{coverage} & 8 & 3 & 776.47 \\
        & & \texttt{NJHillC02} & \texttt{coverage} & 8 & 8 & 783.31 \\
        & & \texttt{NJHillC03} & \texttt{coverage} & 13 & 3 & 761.55 \\
        & & \texttt{NJHillC04} & \texttt{coverage} & 13 & 8 & 768.14 \\
        & & \texttt{NJHillC05} & \texttt{coverage} & 18 & 3 & 756.07 \\
        & & \texttt{NJHillC06} & \texttt{coverage} & 18 & 8 & 769.94 \\
        \midrule
        
        \multirow[c]{11}{*}{Terraced Farmland} & \multirow[l]{11}{*}{\shortstack[l]{Tea plants in regular\\contour rows}} & \texttt{NJTerrB01} & \texttt{boundary} & 3 & 3 & 204.91 \\
        & & \texttt{NJTerrB02} & \texttt{boundary} & 6 & 3 & 207.21 \\
        & & \texttt{NJTerrB03} & \texttt{boundary} & 6 & 6 & 209.71 \\
        & & \texttt{NJTerrB04} & \texttt{boundary} & 9 & 3 & 211.95 \\
        & & \texttt{NJTerrB05} & \texttt{boundary} & 9 & 6 & 215.72 \\
        \cmidrule(lr){3-7}
        & & \texttt{NJTerrC01} & \texttt{coverage} & 3 & 3 & 311.23 \\
        & & \texttt{NJTerrC02} & \texttt{coverage} & 3 & 6 & 307.53 \\
        & & \texttt{NJTerrC03} & \texttt{coverage} & 6 & 3 & 311.24 \\
        & & \texttt{NJTerrC04} & \texttt{coverage} & 6 & 6 & 300.84 \\
        & & \texttt{NJTerrC05} & \texttt{coverage} & 9 & 3 & 313.64 \\
        & & \texttt{NJTerrC06} & \texttt{coverage} & 9 & 6 & 317.48 \\
        \bottomrule
    \end{tabular}
    \label{tab:dataset_sequences}
\end{table*}

\begin{figure*}[t]
    \centering
    \setlength{\tabcolsep}{2pt}
    \renewcommand{\arraystretch}{1.0}

    \begin{tabular}{c c c c}
        & \textbf{Flat} & \textbf{Hilly} & \textbf{Terraced} \\
        \rotatebox{90}{\parbox{4.2cm}{\centering\textbf{Scene}}} &
        \includegraphics[width=0.32\textwidth,height=0.24\textwidth]{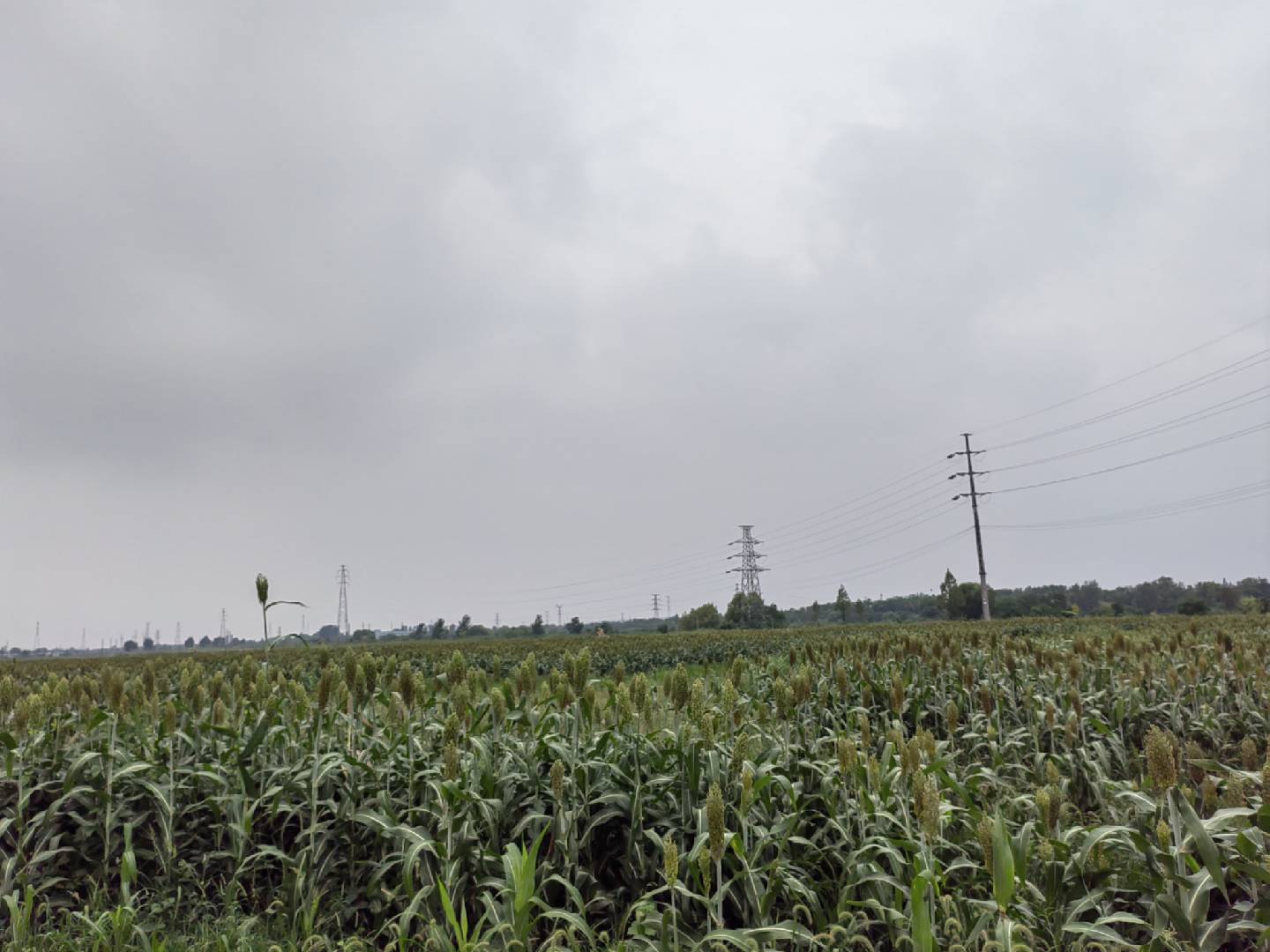} &
        \includegraphics[width=0.32\textwidth,height=0.24\textwidth]{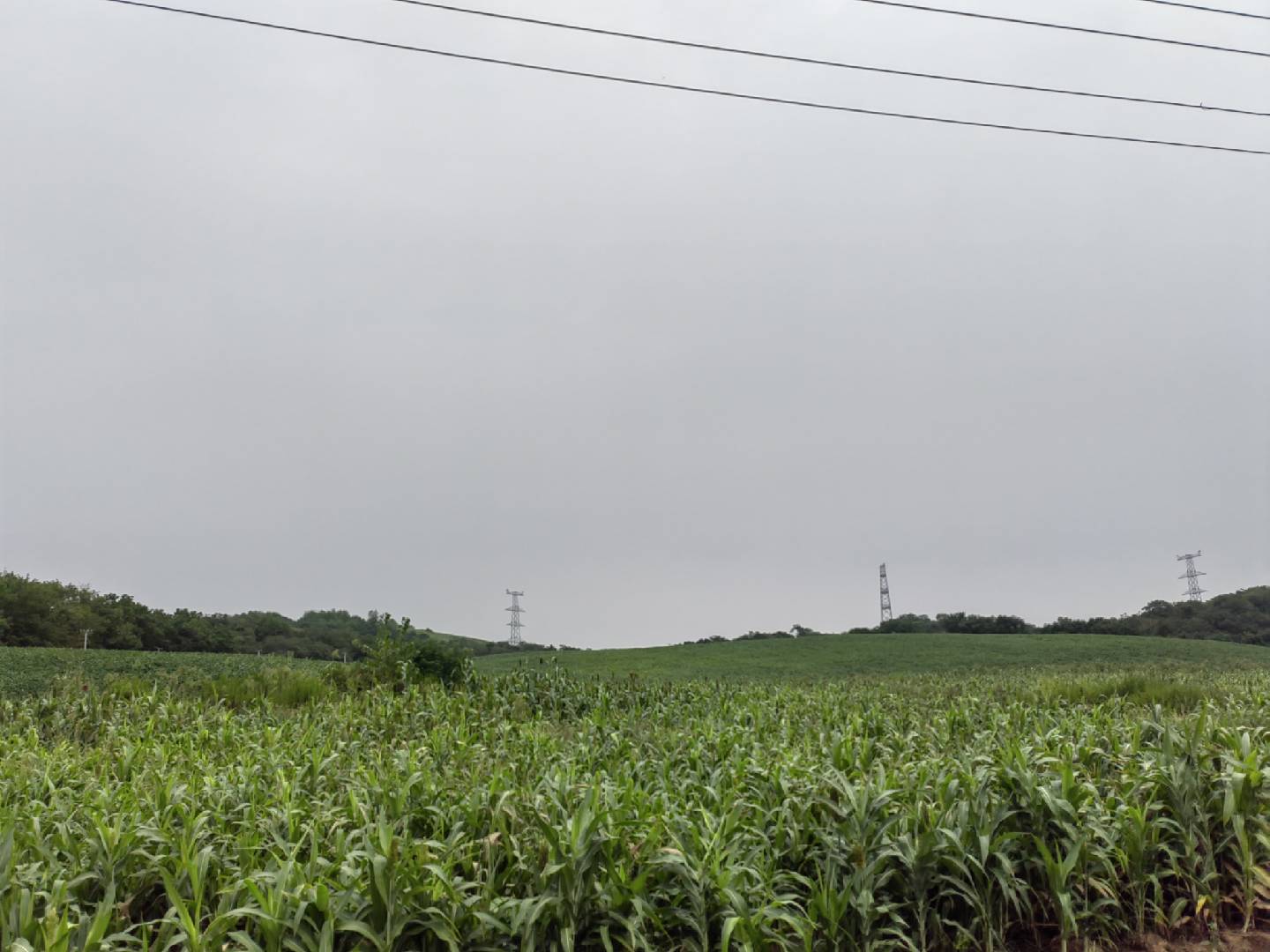} &
        \includegraphics[width=0.32\textwidth,height=0.24\textwidth]{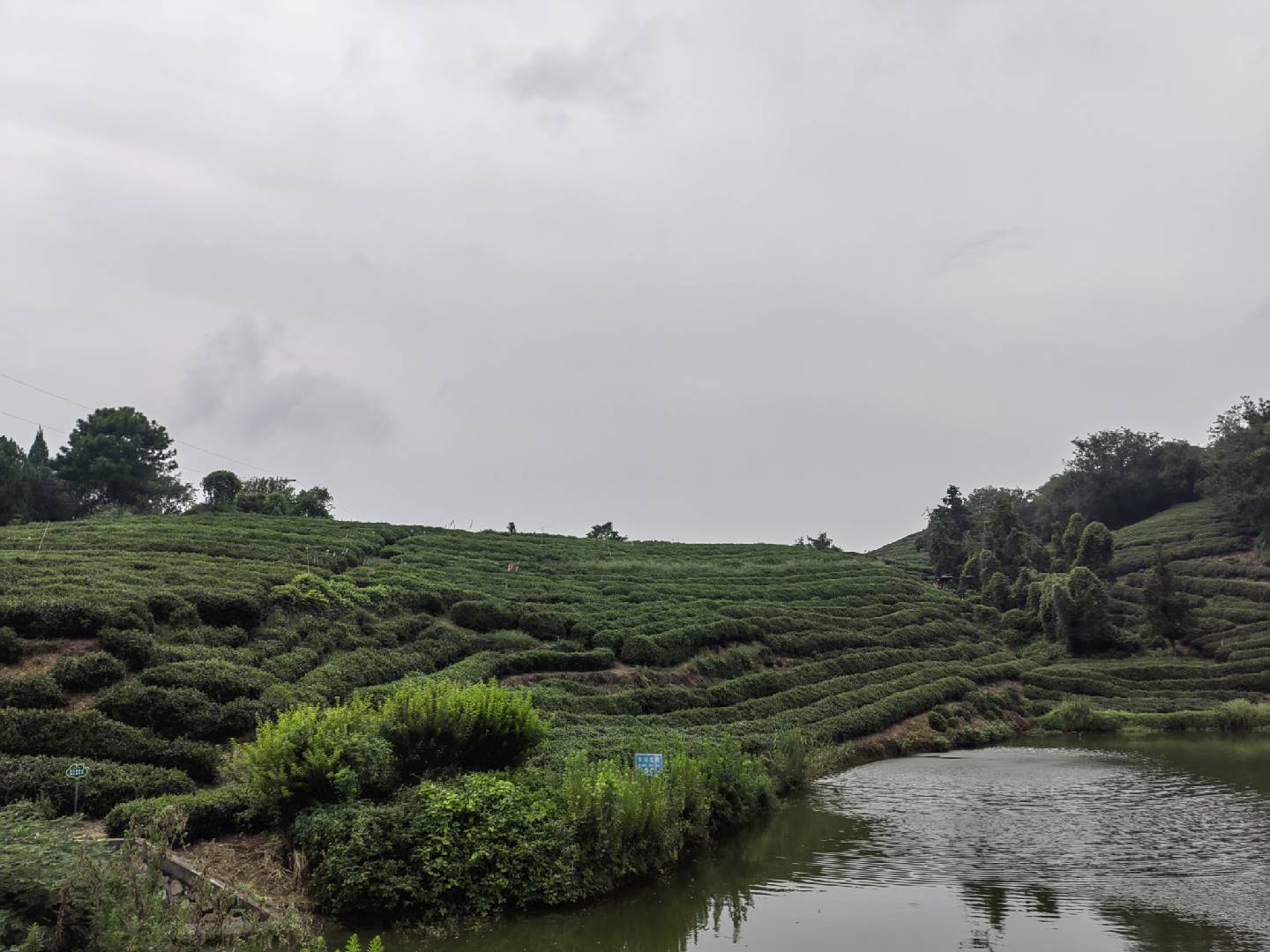} \\
        \rotatebox{90}{\parbox{4.2cm}{\centering\textbf{Scanning Path}}} &
        \includegraphics[width=0.32\textwidth,height=0.24\textwidth]{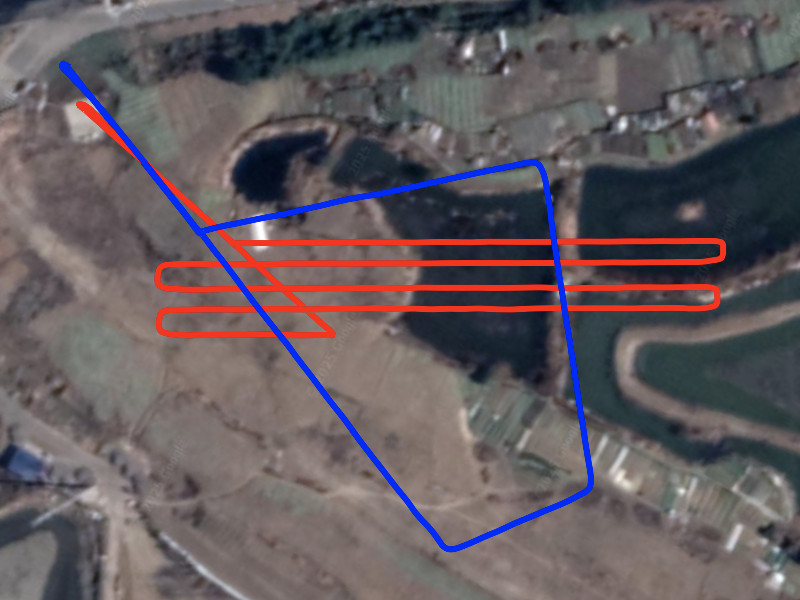} &
        \includegraphics[width=0.32\textwidth,height=0.24\textwidth]{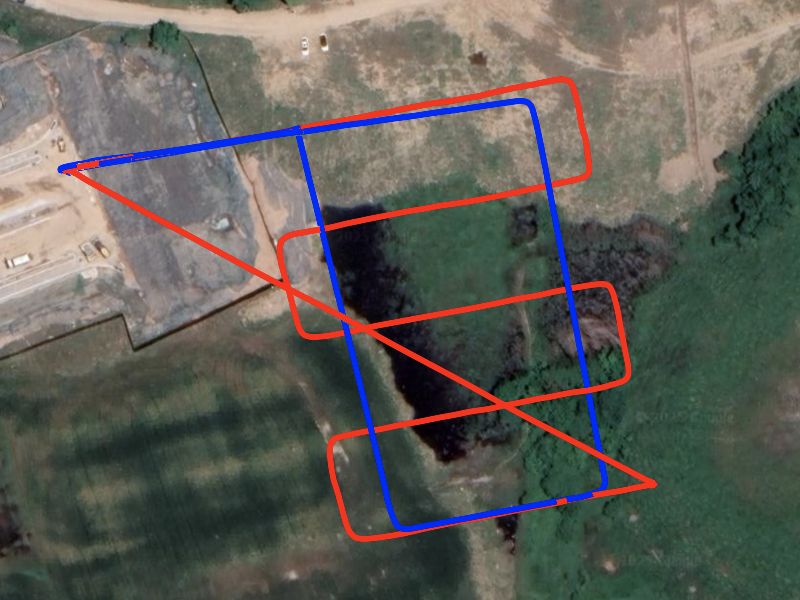} &
        \includegraphics[width=0.32\textwidth,height=0.24\textwidth]{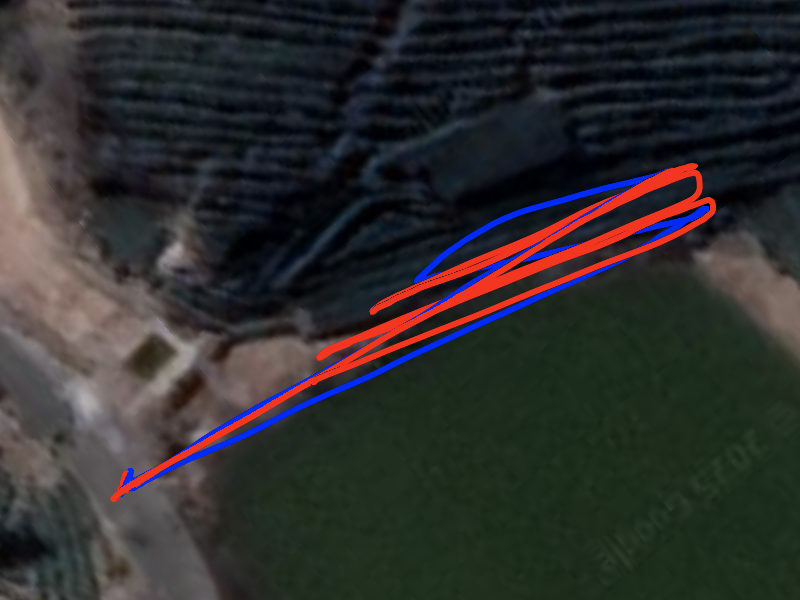} \\
        \rotatebox{90}{\parbox{4.2cm}{\centering\textbf{Lidar}}} &
        \includegraphics[width=0.32\textwidth,height=0.24\textwidth]{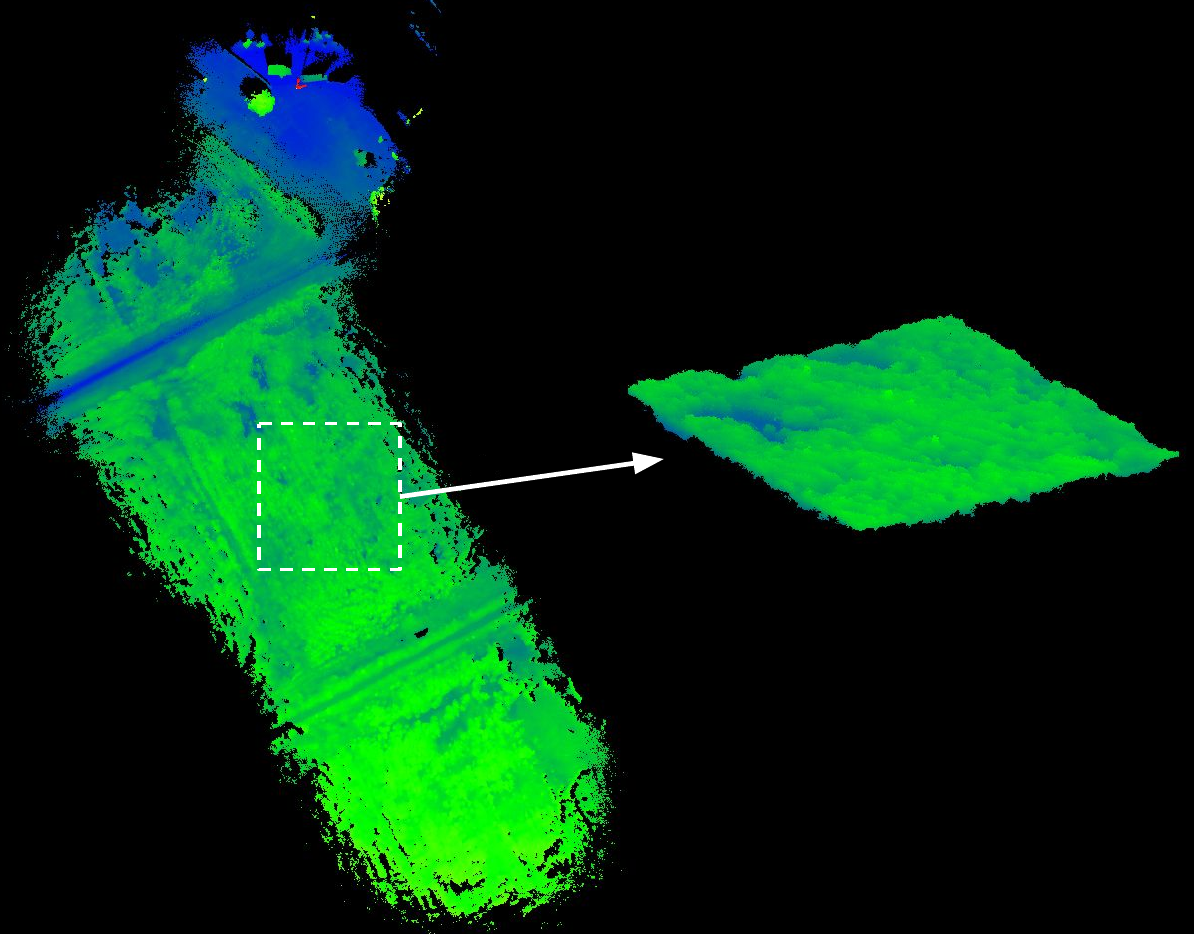} &
        \includegraphics[width=0.32\textwidth,height=0.24\textwidth]{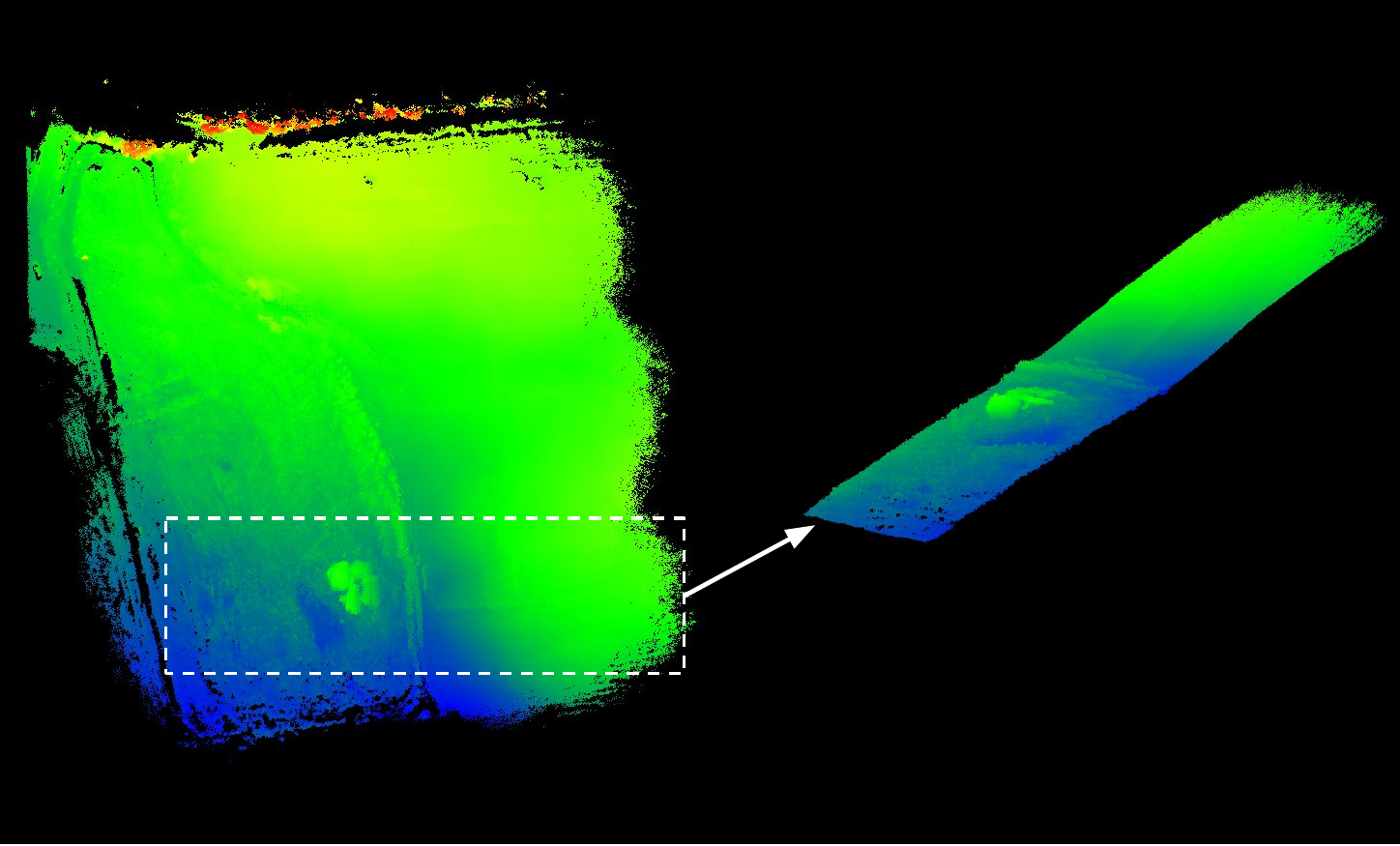} &
        \includegraphics[width=0.32\textwidth,height=0.24\textwidth]{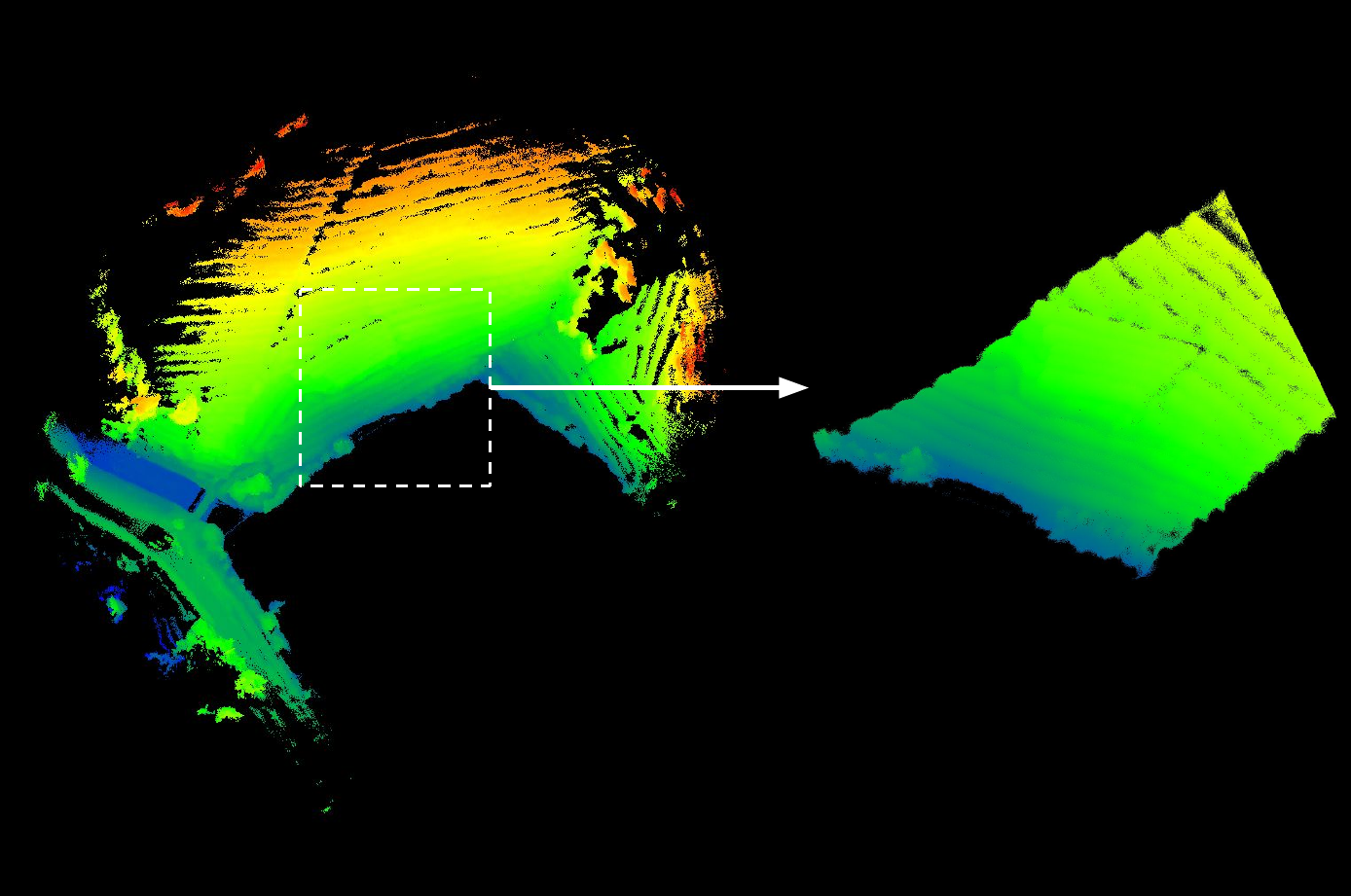} \\
        \rotatebox{90}{\parbox{4.2cm}{\centering\textbf{4D Radar}}} &
        \includegraphics[width=0.32\textwidth,height=0.24\textwidth]{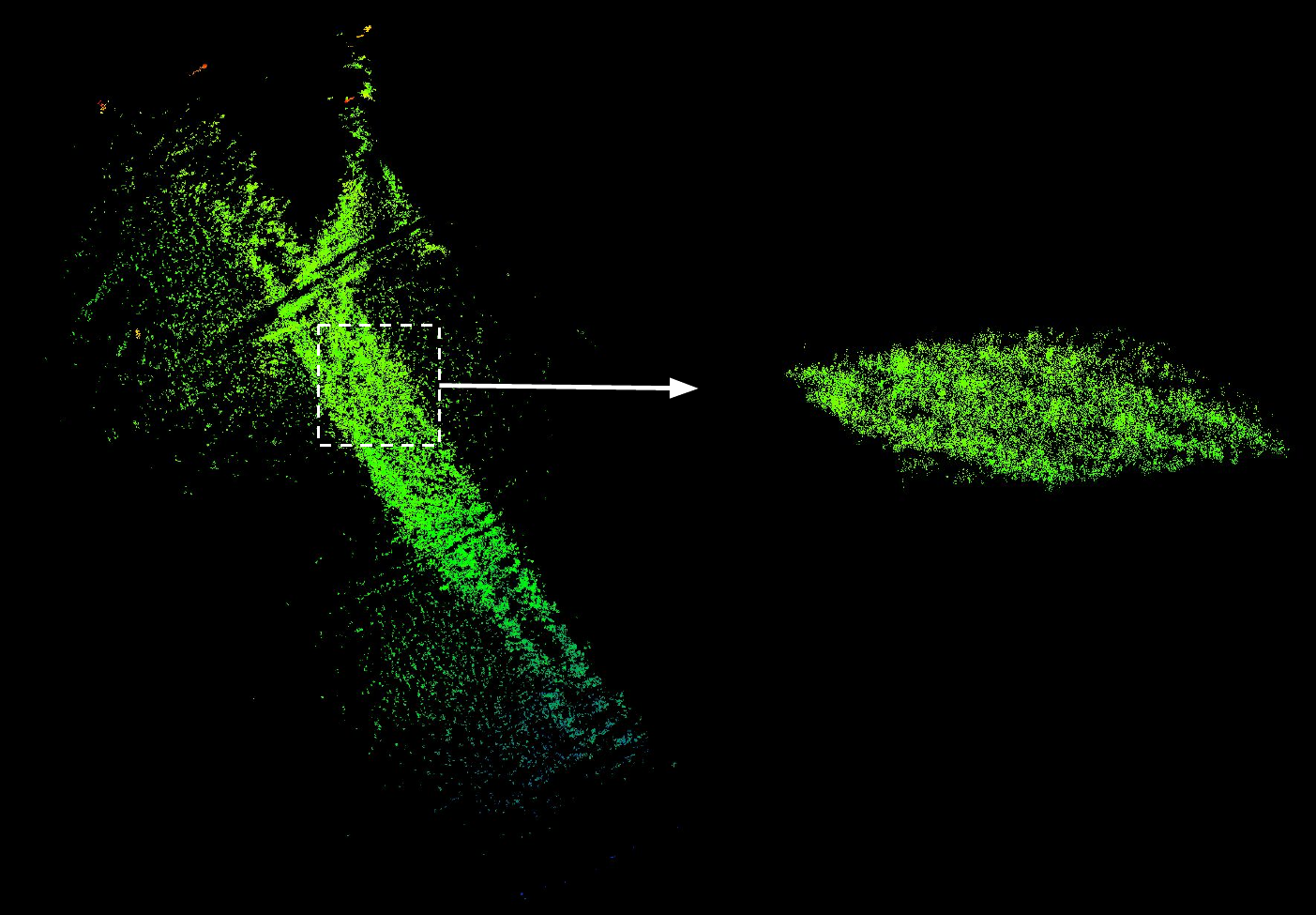} &
        \includegraphics[width=0.32\textwidth,height=0.24\textwidth]{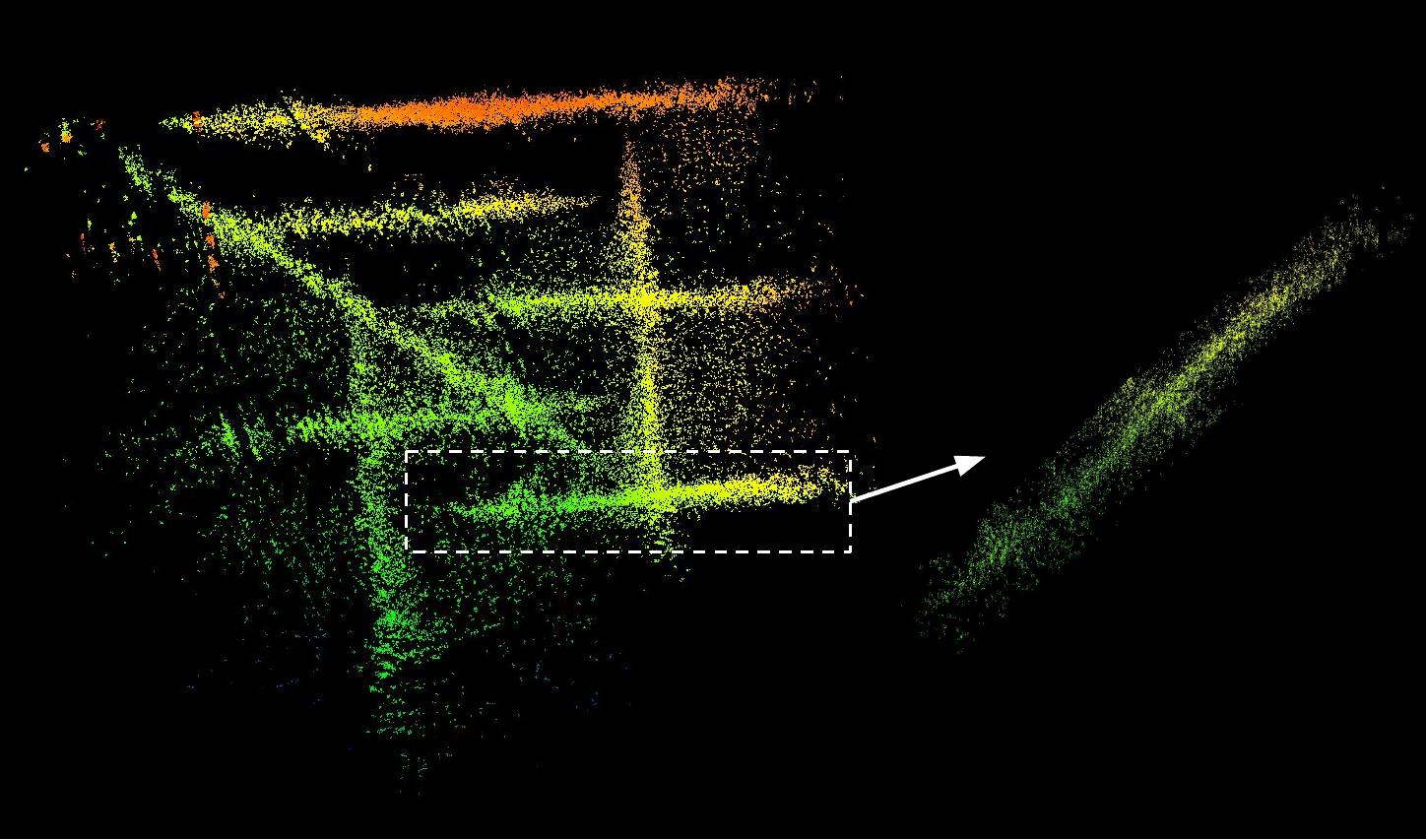} &
        \includegraphics[width=0.32\textwidth,height=0.24\textwidth]{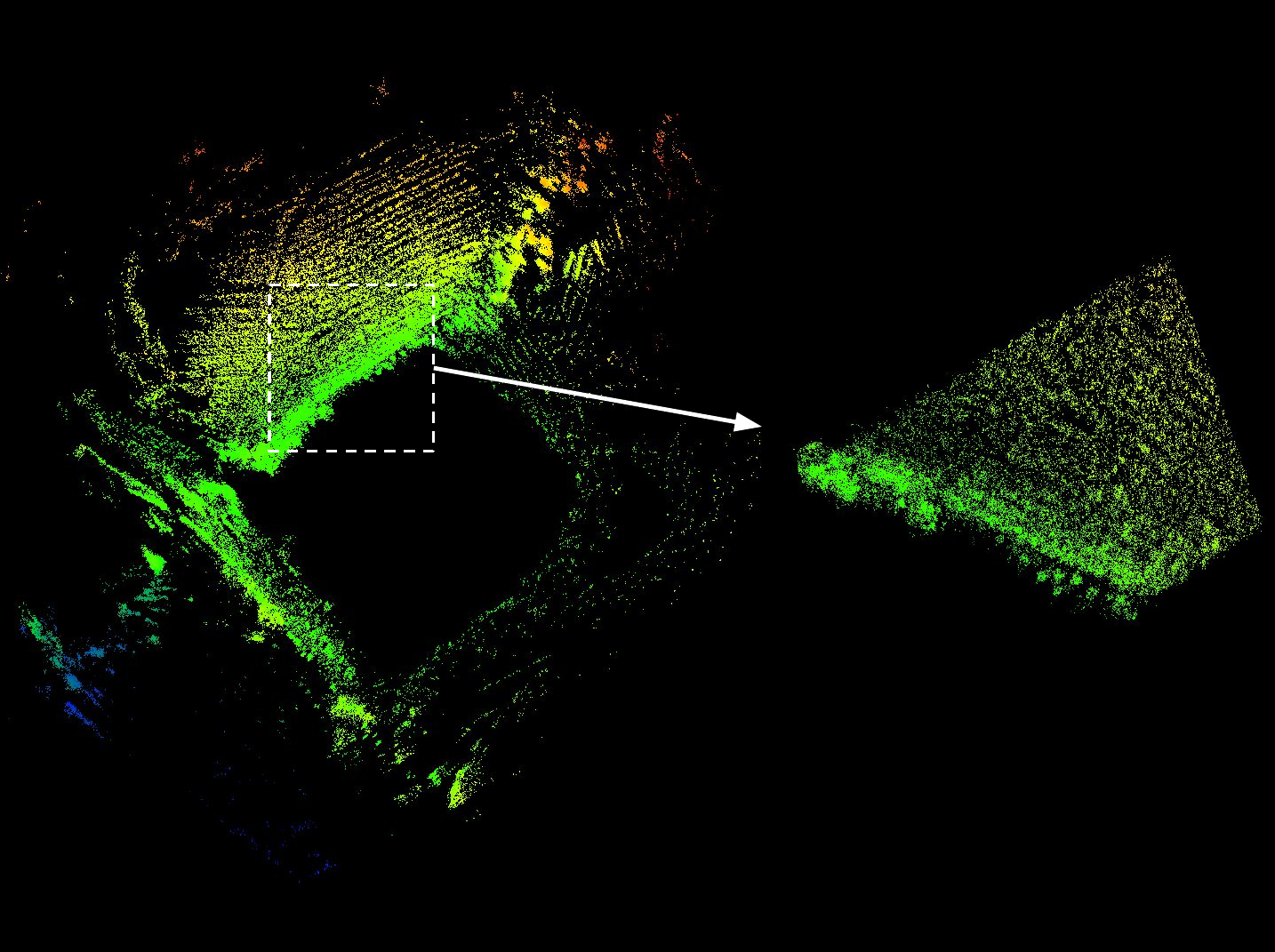} \\
    \end{tabular}
    \caption{\textbf{Visualization of the three representative farmland scenarios and their corresponding sensor data.}
    Each column corresponds to a distinct terrain type: flat farmland, hilly farmland, and terraced farmland. From top to bottom, subfigures illustrate the real-world operation scenes, boundary (\textcolor{blue}{blue}) and coverage (\textcolor{red}{red}) scanning paths, and the top-view height-colored lidar (Faster-LIO~\citep{fasterlio}) and 4D radar (GaRLIO~\citep{noh2025garlio}) maps, with zoomed-in regions highlighting local geometric details.}
    \label{fig:dataset_scenarios}
\end{figure*}

This section characterizes the AgriLiRa4D dataset across three representative farmland terrains—flat plains, hilly regions, and mountainous terraces—located in Nanjing, China. 
The dataset is organized into six sequence groups based on terrain type and scanning mode (\texttt{boundary} or \texttt{coverage}), namely \texttt{NJFlatB}, \texttt{NJFlatC}, \texttt{NJHillB}, \texttt{NJHillC}, \texttt{NJTerrB}, and \texttt{NJTerrC}.
The scope and current limitations of the dataset are further discussed at the end of this section.

The reported flight parameters are predefined target settings, while the actual flight states may vary slightly during operation.
For all sequences except \texttt{NJTerrB} and \texttt{NJTerrC}, the UAV flew at a target altitude with respect to the take-off point.
In contrast, for the mountainous-terrain sequences \texttt{NJTerrB} and \texttt{NJTerrC}, the UAV maintained a target height above ground level (AGL) to ensure flight safety and stable sensor coverage over rapidly varying elevation. 
Multiple combinations of flight altitudes and speeds were employed to introduce different levels of SLAM difficulty. 
Each sequence additionally begins with a short stationary or hovering segment to facilitate IMU initialization.

Table~\ref{tab:dataset_sequences} summarizes the detailed flight configurations and vegetation characteristics for all sequences. 
The real-world operational configurations, along with the corresponding UAV waypoint layouts, are illustrated in Figure~\ref{fig:dataset_scenarios}.

\subsection{Flat Farmland}

The \texttt{NJFlatB} and \texttt{NJFlatC} sequences were collected in flat agricultural fields consisting of nearly mature sorghum awaiting harvest (31.8921°N, 118.8548°E). 
As illustrated in Figure~\ref{fig:dataset_scenarios}, the area spans approximately \qty{250}{\meter} by \qty{350}{\meter}, bordered in part by sparse trees and utility poles. 
The uniformly grown sorghum and largely featureless terrain yield minimal geometric variation, creating a challenging setting for feature extraction, data association, and stable state estimation. 
Two scanning modes were used, with target flight altitudes relative to the take-off point at \qty{5}{\meter}, \qty{10}{\meter}, and \qty{15}{\meter}, and target flight speeds of \qty{3}{\meter\per\second} or \qty{8}{\meter\per\second}, providing sequences with different motion dynamics and viewpoints.

\subsection{Hilly Farmland}

The \texttt{NJHillB} and \texttt{NJHillC} sequences were captured over gently sloped farmland in hilly terrain (31.8348°N, 118.7813°E). The \qty{180}{\meter} by \qty{280}{\meter} area features moderate slopes of approximately \qty{15}{\degree}, with lower regions planted with mature sorghum and upper regions covered by short grass. 
A dense tree line forms one boundary of the site (see Figure~\ref{fig:dataset_scenarios}). 
The mixture of vegetation types and sloped geometry results in more structural cues than flat farmland, though still with limited distinctive features, offering a moderately challenging SLAM environment. 
Data were recorded in two scanning modes at target heights from the take-off point (\qty{8}{\meter}, \qty{13}{\meter}, \qty{18}{\meter}) and at target flight speeds of \qty{3}{\meter\per\second} and \qty{8}{\meter\per\second}.

\subsection{Terraced Farmland}

The \texttt{NJTerrB} and \texttt{NJTerrC} sequences were acquired in steep mountainous terrain consisting of terraced farmland primarily used for tea cultivation (31.7737°N, 118.6917°E). 
The area covers roughly \qty{100}{\meter} by \qty{100}{\meter}, with slopes around \qty{45}{\degree}. 
Tea plants are grown along consistent contour lines, forming distinctive terrace patterns rich in geometric cues, as shown in Figure~\ref{fig:dataset_scenarios}. 
The combination of pronounced elevation changes and structured tea canopies makes this environment relatively favorable for SLAM. 
For safe operation in the steep terrain, the UAV took off from a road at the base of the terraces and maintained a target flight height above ground level (\qty{3}{\meter}, \qty{6}{\meter}, \qty{9}{\meter}), with target flight speeds of \qty{3}{\meter\per\second} or \qty{6}{\meter\per\second}.

\subsection{Dataset Limitations}

AgriLiRa4D dataset spans multiple terrain types, flight configurations, and vegetation structures, but the current release remains geographically concentrated in the Nanjing region and is limited to daytime operation under generally favorable weather. 
Its crop and seasonal coverage is also restricted to a small number of vegetation types and growth conditions. 
Scene-level vegetation characteristics are summarized in Table~\ref{tab:dataset_sequences}; however, quantitative environmental variables, such as wind conditions and vegetation density, were not systematically recorded. 
These factors should be considered when evaluating cross-region, cross-season, and adverse-condition generalization. 
Future extensions will broaden the geographic, seasonal, crop, and environmental coverage and provide more comprehensive scene-level metadata.

\section{Experiments} \label{sec:exp}

\begin{table*}[t]
    \centering
    \caption{\textbf{Translation (meters) and rotation (degrees) RMSE of the ATE for the evaluated methods on the AgriLiRa4D dataset.} 
    For each sequence, results are \textbf{bold} for best, and \underline{underlined} for second best. A dash (“–”) signifies the failure of the algorithm’s execution.}

    \small
    \setlength{\tabcolsep}{9pt}
    \renewcommand{\arraystretch}{1.1}
    
    \begin{tabular}{l l c c c c c c c c}
        \toprule
        \multirow{2}{*}{\textbf{Scene}} & \multirow{2}{*}{\textbf{Sequence}} & \multicolumn{2}{c}{\textbf{FAST-LIO2}} & \multicolumn{2}{c}{\textbf{Faster-LIO}} & \multicolumn{2}{c}{\textbf{EKF-RIO}} & \multicolumn{2}{c}{\textbf{GaRLIO}} \\
        \cmidrule(lr){3-4} \cmidrule(lr){5-6} \cmidrule(lr){7-8} \cmidrule(lr){9-10}
        &  & \textbf{ATE$_{r}$} & \textbf{ATE$_{t}$} & \textbf{ATE$_{r}$} & \textbf{ATE$_{t}$} & \textbf{ATE$_{r}$} & \textbf{ATE$_{t}$} & \textbf{ATE$_{r}$} & \textbf{ATE$_{t}$} \\
        \midrule
        
        \multirow[t]{10}{*}{Flat Farmland} 
        & \texttt{NJFlatB01} & 6.67 & 9.38 & \textbf{4.33} & \textbf{3.73} & - & - & \underline{4.46} & \underline{4.08} \\
        & \texttt{NJFlatB02} & \underline{5.68} & 4.60 & 6.12 & \underline{3.19} & - & - & \textbf{4.35} & \textbf{2.75} \\
        & \texttt{NJFlatB03} & \underline{4.06} & \underline{5.68} & \textbf{3.20} & \textbf{3.76} & - & - & 4.87 & 7.21 \\
        & \texttt{NJFlatB04} & \textbf{3.55} & \underline{3.73} & 3.79 & 4.44 & 19.51 & 29.47 & \underline{3.68} & \textbf{2.91} \\
        & \texttt{NJFlatB05} & 4.57 & 7.04 & \textbf{2.83} & \textbf{3.10} & - & - & \underline{3.02} & \underline{6.07} \\
        & \texttt{NJFlatB06} & \underline{7.64} & \underline{8.69} & \textbf{4.91} & \textbf{4.74} & - & - & 19.82 & 44.65 \\
        \cmidrule(lr){2-10}
        & \texttt{NJFlatC01} & 20.48 & 40.60 & \textbf{6.12} & \textbf{8.37} & \underline{15.93} & \underline{31.56} & 16.80 & 64.36 \\
        & \texttt{NJFlatC02} & 7.15 & 15.05 & \textbf{3.05} & \textbf{4.33} & 16.72 & 82.66 & \underline{5.03} & \underline{7.72} \\
        & \texttt{NJFlatC03} & 4.92 & 7.42 & \textbf{3.08} & \underline{4.54} & 22.67 &	40.39 &	\underline{3.18} & \textbf{4.16} \\
        & \texttt{NJFlatC04} & \underline{3.61} & \underline{5.57} & \textbf{2.80} & \textbf{2.58} & - & - & - & - \\
    
        \midrule    
        
        \multirow[t]{12}{*}{Hilly Farmland} 
        & \texttt{NJHillB01} & \underline{5.16} & \underline{8.05} & \textbf{3.96} & \textbf{4.55} & 6.70 & 11.64 & - & - \\
        & \texttt{NJHillB02} & \underline{4.48} & \underline{4.50} & \textbf{3.78} & \textbf{3.54} & 43.27 & 53.06 & - & - \\
        & \texttt{NJHillB03} & \underline{5.52} & \underline{6.48} & \textbf{4.48} & \textbf{4.24} & 15.77 & 20.33 & - & - \\
        & \texttt{NJHillB04} & \underline{6.58} & \underline{7.85} & \textbf{3.73} & \textbf{4.28} & 43.59 & 56.02 & - & - \\
        & \texttt{NJHillB05} & \textbf{3.60} & 46.15 & 6.08 & 63.22 & 20.60 & \underline{28.18} & \underline{4.43} & \textbf{5.84} \\
        & \texttt{NJHillB06} & \underline{5.30} & \underline{3.80} & \textbf{2.79} & \textbf{3.10} & 14.14 & 28.64 & 7.01 & 25.05 \\
        \cmidrule(lr){2-10}
        & \texttt{NJHillC01} & \underline{15.37} & \underline{47.85} & \textbf{6.46} & \textbf{12.06} & - & - & - & - \\
        & \texttt{NJHillC02} & \underline{8.71} & \underline{18.97} & - & - & - & - & \textbf{7.94} & \textbf{11.69} \\
        & \texttt{NJHillC03} & 6.57 & 41.13 & \underline{6.44} & \underline{20.46} & - & - & \textbf{5.07} & \textbf{10.12} \\
        & \texttt{NJHillC04} & 7.22 & 24.60 & \textbf{3.62} & \underline{5.15} & 9.62 & 23.18 & \underline{4.22} & \textbf{4.36} \\
        & \texttt{NJHillC05} & \textbf{4.52} & \textbf{26.57} & - & - & - & - & - & - \\
        & \texttt{NJHillC06} & \underline{6.29} & \textbf{20.08} & \textbf{3.63} & \underline{26.15} & 42.98 & 64.17 & - & - \\

        \midrule
        
        \multirow[t]{11}{*}{Terraced Farmland} 
        & \texttt{NJTerrB01} & \textbf{2.67} & \underline{0.70} & \underline{2.69} & \textbf{0.67} & 5.42 & 5.03 & 3.06 & 1.67 \\
        & \texttt{NJTerrB02} & \textbf{2.61} & \textbf{0.51} & \underline{2.64} & \underline{0.61} & 17.59 & 9.33 & 21.54 & 15.41 \\
        & \texttt{NJTerrB03} & \textbf{2.22} & \underline{0.51} & \underline{2.55} & \textbf{0.50} & 32.92 & 15.91 & 21.18 & 17.62 \\
        & \texttt{NJTerrB04} & \textbf{2.63} & \underline{0.74} & \underline{2.77} & \textbf{0.73} & 3.43 & 4.03 & 38.13 & 17.82 \\
        & \texttt{NJTerrB05} & \underline{6.60} & \underline{3.60} & \textbf{3.70} & \textbf{1.50} & - & - & 20.56 & 14.50 \\
        \cmidrule(lr){2-10}
        & \texttt{NJTerrC01} & \textbf{3.02} & \underline{0.93} & \underline{3.07} & \textbf{0.92} & - & - & 8.35 & 5.12 \\
        & \texttt{NJTerrC02} & \underline{4.39} & \underline{2.27} & \textbf{2.79} & \textbf{0.76} & 21.93 & 13.49 & 12.98 & 7.66 \\
        & \texttt{NJTerrC03} & \textbf{2.63} & \textbf{1.14} & \underline{3.56} & \underline{2.47} & 12.20 & 7.24 & 7.16 & 11.54 \\
        & \texttt{NJTerrC04} & \underline{2.67} & \textbf{1.21} & \textbf{2.66} & \underline{1.66} & - & - & 8.14 & 10.46 \\
        & \texttt{NJTerrC05} & 2.32 & \underline{1.02} & \textbf{2.22} & \textbf{0.62} & 6.14 & 6.72 & \underline{2.27} & 1.05 \\
        & \texttt{NJTerrC06} & \textbf{2.70} & \textbf{1.89} & 25.21 & 21.58 & - & - & \underline{4.05} & \underline{2.82} \\

        \midrule
        
        \multirow[t]{3}{*}{Overall} 
        & \texttt{Mean} & \underline{5.52} & \underline{11.46} & \textbf{4.49} & \textbf{7.15} & 19.53 & 27.95 & 9.65 & 12.27  \\
        & \texttt{Median} & \underline{4.57} & \underline{5.68} & \textbf{3.62} & \textbf{3.73} & 16.72 & 23.18 & 5.07 & 7.66 \\
        & \texttt{Success rate} & \multicolumn{2}{c}{\textbf{100\%}} & \multicolumn{2}{c}{\underline{93.9\%}} & \multicolumn{2}{c}{57.6\%} & \multicolumn{2}{c}{75.8\%}  \\
        \bottomrule
    \end{tabular}
    \label{tab:evo_result}
\end{table*}

\begin{table*}[t]
\centering
\caption{\textbf{Success rates of the evaluated methods grouped by terrain, relative flight altitude, nominal flight speed, and scanning mode.}
The number of sequences in each group is denoted by ``Sequences''.}
\label{tab:aggregate_success_rates}

\small
\setlength{\tabcolsep}{10pt}
\renewcommand{\arraystretch}{1.1}

\begin{tabular}{llccccc}
\toprule
\textbf{Category}
& \textbf{Group}
& \textbf{Sequences}
& \textbf{FAST-LIO2}
& \textbf{Faster-LIO}
& \textbf{EKF-RIO}
& \textbf{GaRLIO} \\
\midrule

\multirow{3}{*}{Terrain}
& Flat
& 10
& 100.0\%
& 100.0\%
& 40.0\%
& 90.0\% \\

& Hilly
& 12
& 100.0\%
& 83.3\%
& 66.7\%
& 41.7\% \\

& Terraced
& 11
& 100.0\%
& 100.0\%
& 63.6\%
& 100.0\% \\

\midrule

\multirow{3}{*}{Altitude}
& High
& 11
& 100.0\%
& 90.9\%
& 45.5\%
& 72.7\% \\

& Medium
& 12
& 100.0\%
& 100.0\%
& 75.0\%
& 83.3\% \\

& Low
& 10
& 100.0\%
& 90.0\%
& 50.0\%
& 70.0\% \\

\midrule

\multirow{3}{*}{Speed}
& 3\,m/s
& 17
& 100.0\%
& 94.1\%
& 52.9\%
& 70.6\% \\

& 6\,m/s
& 5
& 100.0\%
& 100.0\%
& 40.0\%
& 100.0\% \\

& 8\,m/s
& 11
& 100.0\%
& 90.9\%
& 72.7\%
& 72.7\% \\

\midrule

\multirow{2}{*}{Scanning Mode}
& Boundary
& 17
& 100.0\%
& 100.0\%
& 64.7\%
& 76.5\% \\

& Coverage
& 16
& 100.0\%
& 87.5\%
& 50.0\%
& 75.0\% \\

\bottomrule
\end{tabular}

\vspace{2pt}
\begin{minipage}{0.98\textwidth}
\footnotesize
\textit{Note:}
For the altitude-based grouping, low, medium, and high denote the three terrain-specific flight levels: 5, 10, and 15\,m for flat farmland; 8, 13, and 18\,m for hilly farmland; and 3, 6, and 9\,m AGL for terraced farmland, respectively.
The speed groups are not uniformly distributed across terrain types: 6\,m/s occurs only in terraced farmland, whereas 8\,m/s occurs only in flat and hilly farmland. 
Therefore, the grouped results are descriptive and should not be interpreted as isolating the causal effect of flight speed.
\end{minipage}
\end{table*}

This section presents the benchmark evaluation of AgriLiRa4D. 
We first describe the evaluated methods and experimental protocol, and then examine overall localization performance, sensing-modality effects, and the influence of terrain, vegetation, and flight altitude.

\subsection{Benchmark Methods and Evaluation}
To thoroughly assess the proposed AgriLiRa4D dataset, we benchmark representative state-of-the-art multi-sensor SLAM algorithms covering three major sensing modalities: lidar-inertial odometry (LIO), radar-inertial odometry (RIO), and radar-lidar-inertial odometry (RLIO).
LIO approaches such as FAST-LIO2~\citep{fastlio2}, Faster-LIO~\citep{fasterlio}, LIO-SAM~\citep{liosam2020shan}, and LiLi-OM~\citep{liliom} fuse dense lidar scans with IMU integration and are widely adopted for precise odometry in structured and semi-structured environments.
RIO methods, including EKF-RIO~\citep{DoerENC2020}, DRIO~\citep{drio}, and 4D-IRIOM~\citep{zhuang20234d}, leverage radar’s robustness against visual and geometric degradation, while doppler measurements provide additional motion cues when lidar becomes unreliable.
RLIO frameworks such as DR-LRIO~\citep{drrlio}, AF-RLIO~\citep{AFRLIO}, and GaRLIO~\citep{noh2025garlio} further integrate the complementary characteristics of all three modalities to achieve enhanced consistency and robustness under challenging conditions.

Considering practical constraints such as open-source availability, implementation stability, and compatibility with our GNSS-free and vision-free evaluation protocol, we selected four algorithms for benchmarking: FAST-LIO2 and Faster-LIO for LIO, EKF-RIO for RIO, and GaRLIO for RLIO. Other representative methods discussed above were not included because many lack official open-source implementations, rely on sensing modalities that are not available in our dataset (such as visual, depth, or GNSS measurements), or require hardware interfaces and engineering adaptations that hinder reproducible large-scale evaluation. 
Overall, the selected four algorithms provide representative and reliable baselines that cover the LIO, RIO, and RLIO modality spectrum for benchmarking on AgriLiRa4D.

For reproducibility, all four methods are evaluated using the \texttt{agrilira4d} branches in our forked repositories of FAST-LIO2, Faster-LIO, EKF-RIO, and GaRLIO. 
These branches contain only the necessary adaptations for the AgriLiRa4D sensor interfaces, while the original upstream parameter settings are retained without method-specific tuning. 
All experiments are conducted on a desktop computer equipped with an Intel Core i7-12700 CPU and 32~GB RAM, running Ubuntu~20.04 and ROS Noetic.

A common IMU-based initialization procedure is used for all methods, where the initial stationary/hovering measurements are used to estimate the gravity direction. 
The initialization interval is retained in the quantitative evaluation to account for initialization accuracy.
Performance is evaluated using the root mean square error (RMSE) of the absolute trajectory error (ATE), computed by directly comparing each estimated trajectory with the high-precision FINS-RTK ground truth in the released FLU evaluation frame.
All evaluations are performed using the evo toolkit~\citep{grupp2017evo} to ensure consistency and reproducibility, with translational and rotational errors reported in meters and degrees, respectively.
No post-hoc alignment or scale correction is applied in evo, since additional alignment would partially absorb initialization-related pose errors and all evaluated methods already produce metric-scale trajectories.

The quantitative results across different farmland scenarios and flight configurations are summarized in Table~\ref{tab:evo_result}, providing a comprehensive comparison of the selected algorithms under varied environmental and operational conditions. A run is considered a failure if the translational ATE relative to the path length (Table~\ref{tab:dataset_sequences}) exceeds 15\%, or if the rotational ATE exceeds \qty{45}{\degree}.
These thresholds were empirically selected based on inspection of the evaluated trajectories, where they provided a practical separation between clear divergence and large but still interpretable drift. 
The translational error is normalized by path length to account for differences in sequence scale, while the rotational threshold captures severe orientation failure. 
These criteria are applied consistently across all methods and sequences and are not intended as universal SLAM failure thresholds.

\subsection{Overall Benchmark Performance}

Across all evaluated sequences, the LIO methods deliver the most stable and consistently accurate performance on the AgriLiRa4D dataset. 
Both FAST-LIO2 and Faster-LIO achieve the lowest ATE in the majority of sequences and maintain reliable operation across all farmland types, with only moderate degradation under higher altitudes or faster flight speeds. 
In contrast, the RIO baseline, EKF-RIO, exhibits the least stable behavior, failing on nearly all flat farmland sequences and showing large drift even in those hilly and terraced sequences where it successfully completes a run. 
The RLIO method GaRLIO demonstrates mixed performance: it frequently fails in hilly farmland, operates inconsistently in flat terrain, but achieves its most stable and competitive results in terraced farmland, where it occasionally matches or slightly outperforms the LIO baselines in rotation accuracy.
Overall, the benchmark outcomes highlight a wide difficulty spectrum across sensing modalities and farmland types, demonstrating that AgriLiRa4D offers a comprehensive and discriminative testbed for evaluating multi-sensor SLAM robustness in real agricultural environments.

The aggregate statistics in Table~\ref{tab:evo_result} provide a more concise comparison of the evaluated methods.
FAST-LIO2 completes all 33 sequences, while Faster-LIO, EKF-RIO, and GaRLIO achieve success rates of 93.9\%, 57.6\%, and 75.8\%, respectively. 
Among successful runs, Faster-LIO achieves the lowest mean and median translational ATE, at 7.15\,m and 3.73\,m, respectively. 
FAST-LIO2 achieves corresponding values of 11.46\,m and 5.68\,m, while GaRLIO achieves 12.27\,m and 7.66\,m. 
EKF-RIO exhibits the largest aggregate errors, with a mean and median translational ATE of 27.95\,m and 23.18\,m. 
The rotational ATE statistics exhibit a similar overall trend. 
These results confirm that the two LIO methods provide the most consistent performance, whereas the radar-based methods exhibit greater variability across sequences.

\subsection{Impact of Sensing Modality}

The performance differences observed among LIO, RIO, and RLIO methods can be largely attributed to the sensing characteristics and the original design intentions of each modality. 
LIO approaches, represented by FAST-LIO2 and Faster-LIO, inherently benefit from dense geometric constraints and tight lidar–IMU coupling. Both methods were originally validated on UAV platforms—including indoor and outdoor flight scenarios—making them naturally suited for the fast motion, wide-area scanning, and geometry-rich conditions typical of agricultural UAV operation. This strong geometric anchoring explains why LIO methods remain the most stable across all terrain types in our dataset.

In contrast, the RIO baseline EKF-RIO shows limited robustness in outdoor agricultural environments. 
Although EKF-RIO was originally evaluated on UAVs, its validation was restricted to indoor settings, where radar returns are much denser and more structured. In large open agricultural fields, 4D radar often produces sparse and noisy returns, especially over flat crops or grass surfaces, resulting in insufficient geometric constraints for stable odometry. 
Moreover, radar measurements are highly sensitive to vegetation motion induced by UAV downwash, yielding returns that shift unpredictably across different parts of the plant canopy—and occasionally even from the ground surface. 
Such inconsistent and spatially drifting reflections introduce severe ambiguity, thus making stable data association particularly challenging. 
While doppler velocity provides valuable motion cues, these cues become unreliable when the underlying spatial structure is weak, contributing to the significant drift and frequent failures observed in our evaluation.

\begin{figure}[htbp]
    \centering
    \includegraphics[width=0.5\textwidth]{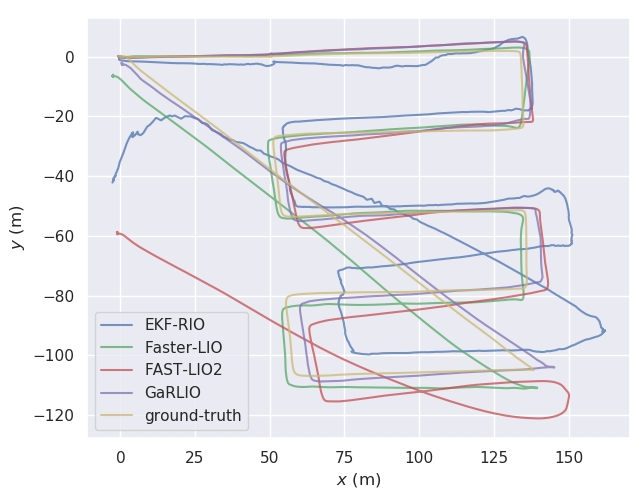}
    \caption{\textbf{Top-down view comparison of trajectories for the \texttt{NJHillC04} sequence.} 
    The FINS-RTK ground truth trajectory is presented as the benchmark reference, together with four algorithm-estimated trajectories for performance comparison.}
    \label{fig:traj_xy}
\end{figure}

\begin{figure}[htbp]
    \centering
    \includegraphics[width=0.5\textwidth]{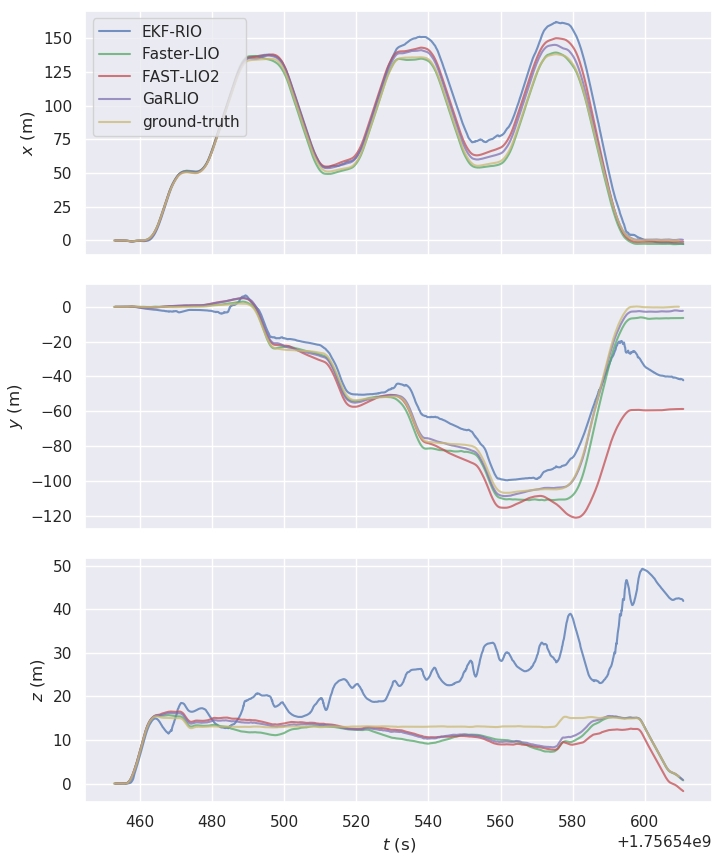}
    \caption{\textbf{Position errors for the \texttt{NJHillC04} sequence.} 
    The errors of four algorithms relative to the FINS-RTK ground truth are decomposed and displayed along the X, Y, and Z axes, respectively.}
    \label{fig:pos_error_xyz}
\end{figure}

The RLIO method GaRLIO exhibits mixed performance due to its sensing design and domain mismatch. 
GaRLIO was originally developed and tested primarily on unmanned ground vehicle (UGV) platforms with near-ground viewpoints, balanced lidar-radar overlap, and slower, more stable motion profiles. 
When applied to UAV flights, several mismatches arise: lidar-radar common visibility decreases with altitude, doppler signatures change under fast aerial motion, and radar becomes more sensitive to vegetation-driven perturbations. 
Consequently, GaRLIO occasionally outperforms LIO in hilly farmland as shown in Figure~\ref{fig:traj_xy} and Figure~\ref{fig:pos_error_xyz}, where the sloped terrain provides stronger and more distinguishable radar returns that compensate for its reduced lidar support.

The sequence-level results nevertheless reveal cases in which radar provides useful complementary constraints. 
For example, on \texttt{NJHillB05}, GaRLIO achieves a translational ATE of 5.84\,m, compared with 46.15\,m for FAST-LIO2 and 63.22\,m for Faster-LIO.
Similar behavior is observed on \texttt{NJHillC03} and \texttt{NJHillC04}. 
In these cases, the longer sensing range of radar can capture the sloped terrain beyond the locally observed lidar region, while doppler-derived radial-velocity measurements can complement locally weak lidar constraints.

However, the complementary benefit of radar is not consistent across all sequences. 
A representative failure case is \texttt{NJHillC05}, a coverage sequence collected at a target altitude of 18\,m, in which only FAST-LIO2 completes the trajectory. 
The hilly site contains both mature sorghum and extensive short-grass regions.
Compared with taller sorghum, short grass forms a more homogeneous and locally planar surface from the airborne lidar viewpoint and therefore provides fewer distinctive geometric constraints. 
The higher altitude further reduces lidar point density and lidar-radar overlap.
Meanwhile, sparse radar returns and vegetation motion induced by wind or UAV downwash can produce unstable spatial and doppler measurements.
These combined effects likely contribute to the failures observed in this sequence. 
Another representative case is \texttt{NJFlatC04}, in which EKF-RIO and GaRLIO fail while both LIO methods remain operational. 
Although the taller sorghum provides more three-dimensional structure than short grass, its repetitive arrangement and vegetation motion can still produce ambiguous geometric associations and unstable radar returns during a high-altitude, long-distance coverage flight.

Across all modalities, a common trend is that positional drift predominantly accumulates along the vertical (Z) axis, as illustrated in Figure~\ref{fig:pos_error_xyz} for the \texttt{NJHillC04} sequence. 
This behavior is consistent with the sensing geometry: airborne lidar primarily observes the environment from near-horizontal viewpoints, providing weaker constraints on altitude compared to horizontal motion, while 4D radar exhibits coarser elevation resolution. 
As a result, Z-axis estimation relies more heavily on IMU integration, making it particularly susceptible to drift when geometric structure or radar elevation cues are insufficient.

These results should, however, be interpreted carefully.
Within the evaluated baselines, the lidar-inertial methods provide the most consistent overall performance, whereas the effectiveness of the radar-based and radar-lidar-inertial methods varies substantially across terrains and flight conditions. 
Moreover, because only one RLIO baseline is evaluated and no controlled sensor-ablation study is conducted, the results primarily characterize the behavior of the selected implementations rather than the upper-bound potential of multi-modal fusion.

\subsection{Effect of Terrain, Vegetation and Altitude}

The varying performance across different farmland types in AgriLiRa4D can be explained by the combined effects of terrain geometry, vegetation characteristics, and flight parameters such as altitude and speed. 
As shown in Figure~\ref{fig:dataset_scenarios}, terraced farmland consistently yields the most favorable results for all methods due to its strong and well-structured geometric contours. 
The steep slopes and layered tea canopies create stable depth gradients that offer abundant geometric constraints for lidar-based matching and informative doppler patterns for radar processing. 
These structural advantages allow even the weaker sensing modalities, such as RIO, to complete trajectories reliably and also lead to GaRLIO’s best overall performance in this terrain category.

In contrast, flat farmland represents the most challenging scenario in the dataset. 
The nearly uniform sorghum fields provide minimal 3D geometric variation, resulting in sparse lidar features and weak radar returns, illustrated in Figure~\ref{fig:dataset_scenarios}. 
The highly repetitive crop-row patterns further complicate data association, magnifying drift for all SLAM systems. 
Dynamic vegetation motion induced by UAV downwash contributes additional instability: sorghum stalks sway noticeably at the evaluated flight speeds of \qty{3}{\meter\per\second} and \qty{8}{\meter\per\second}, generating fluctuating lidar edges and volatile radar reflections. 
These factors collectively explain the significant performance degradation observed in flat farmland, particularly in the high-speed (\texttt{NJFlatB02/B04/B06}) and long-path coverage sequences (\texttt{NJFlatC01-C04}), where the longest trajectories exceed \qty{800}{\meter}.

Hilly farmland presents intermediate difficulty. 
Although the terrain contains noticeable elevation changes, the local surface patches observed from typical UAV altitudes remain nearly planar for lidar, offering limited improvement for LIO compared with flat fields. 
In contrast, radar is more sensitive to terrain gradients, primarily due to its substantially longer sensing range. 
This extended visibility enables RLIO to occasionally benefit from the sloped ground geometry. 
At the same time, grassy and shrub-covered surfaces in hilly regions remain vulnerable to wind-induced motion, introducing fluctuations in both lidar and radar measurements. 

Flight altitude further modulates the effective sensing quality across all scenes. 
As altitude increases from \qty{3}{\meter} to \qty{18}{\meter} (see Table~\ref{tab:dataset_sequences}), lidar beams intersect the ground at progressively shallower angles, reducing point density and diminishing geometric distinctiveness. 
Radar detections also become sparser and less reliable at higher elevations. 
These altitude-induced limitations also reduce the spatial overlap between lidar and radar, weakening the cross-modal constraints required by RLIO systems. 
As a result, sequences flown at higher altitudes—such as \texttt{NJFlatC04} and \texttt{NJHillC05}—exhibit higher failure rates and larger trajectory errors across all modalities.

The grouped success rates in Table~\ref{tab:aggregate_success_rates} further summarize these environmental and operational trends. 
GaRLIO achieves success rates of 90.0\% and 100.0\% in flat and terraced farmland, respectively, but only 41.7\% in hilly farmland, reflecting its strong sensitivity to the availability and stability of cross-modal constraints.
EKF-RIO achieves its lowest terrain-specific success rate in flat farmland, at 40.0\%. Coverage sequences are also more challenging than boundary sequences for Faster-LIO and EKF-RIO, whose success rates decrease from 100.0\% to 87.5\% and from 64.7\% to 50.0\%, respectively. 
No monotonic relationship is observed between success rate and either altitude or nominal flight speed. 
Because the altitude and speed settings differ across terrain types and do not form a fully balanced factorial design, these grouped results are interpreted descriptively rather than as estimates of the independent effects of individual flight parameters.


\section{Conclusion and Future Work} \label{sec:conclusion}

In this paper, we present AgriLiRa4D, a multi-modal UAV dataset specifically designed to address the challenges of SLAM and localization in real agricultural environments. 
The dataset covers three representative farmland types and two flight modes, delivering diverse motion patterns and sensing conditions that are rarely included in existing benchmarks. 
AgriLiRa4D offers high-precision FINS-RTK ground truth, time-synchronized lidar, 4D radar, and IMU measurements, as well as full calibration data, providing a reliable foundation for developing and evaluating multi-sensor fusion algorithms. 
Using this dataset, we benchmark several state-of-the-art multi-sensor SLAM algorithms and highlight the intrinsic difficulties posed by low-texture crops, repetitive planting structures, uneven terrain, and vegetation dynamics. 
These results underscore the value of AgriLiRa4D for studying the strengths, limitations, and operating conditions of different sensing modalities and for developing more effective multi-modal fusion methods.

Looking ahead, we plan to extend AgriLiRa4D with additional crop types, seasonal variations, and more extreme operational conditions, such as night flights and adverse weather, to further support research on resilient agricultural autonomy. 
We believe AgriLiRa4D will serve as a long-term resource for the robotics community, catalyzing progress in robust perception and navigation for agricultural UAVs. 
The challenges surfaced through this benchmark will guide our future work toward developing more resilient, adaptive, and generalizable SLAM frameworks capable of handling the complex, dynamic, and often ambiguous conditions intrinsic to real-world agricultural environments.


\begin{acks}
 The authors thank TopXGun Robotics for supporting field experiments and system deployment.
\end{acks}

\begin{funding}
This research has been supported in part by the National Natural Science Foundation of China under Grant 62401188 and Zhejiang Provincial Natural Science Foundation of China under Grant LQN25F030015.
\end{funding}

\bibliographystyle{SageH}
\bibliography{ref}

\end{document}